\pdfoutput=1
\documentclass[runningheads]{llncs}
\usepackage{graphicx}
\usepackage{amssymb}
\usepackage{amsfonts}
\usepackage{amsmath}
\usepackage{bm}
\usepackage{subfig}
\usepackage{booktabs} 
\usepackage{algorithm}
\usepackage{algpseudocode}
\usepackage{bigstrut,multirow,diagbox}
\usepackage{hyperref}
\usepackage{color}
\usepackage{cite}
\begin{document}
\renewcommand{\algorithmicrequire}{\textbf{Input:}}
\renewcommand{\algorithmicensure}{\textbf{Output:}}
\title{Self-Paced Probabilistic Principal Component Analysis for Data with Outliers}
\titlerunning{Self-Paced Probabilistic PCA}
%

\author{Bowen Zhao\inst{1} \and
Xi Xiao\inst{1,}\inst{2} \and
Wanpeng Zhang\inst{1} \and
Bin Zhang\inst{2} \and
Shutao Xia\inst{1,}\inst{2}}
\authorrunning{B.W. Zhao et al.}
%
\institute{Graduate School At Shenzhen, Tsinghua University, Shenzhen, China\\
\email{zbw18@mails.tsinghua.edu.cn, xiaox@sz.tsinghua.edu.cn\\ zawnpn@gmail.com, xiast@sz.tsinghua.edu.cn}
\and
Peng Cheng Laboratory, Shenzhen, China\\
\email{bin.zhang@pcl.ac.cn}
}
%

\maketitle              
\begin{abstract}
Principal Component Analysis (PCA) is a popular tool for dimensionality reduction and feature extraction in data analysis. There is a probabilistic version of PCA, known as Probabilistic PCA (PPCA). However, standard PCA and PPCA are not robust, as they are sensitive to outliers. To alleviate this problem, this paper introduces the Self-Paced Learning mechanism into PPCA, and proposes a novel method called Self-Paced Probabilistic Principal Component Analysis (SP-PPCA). Furthermore, we design the corresponding optimization algorithm based on the alternative search strategy and the expectation-maximization algorithm. SP-PPCA looks for optimal projection vectors and filters out outliers iteratively. Experiments on both synthetic problems and real-world datasets clearly demonstrate that SP-PPCA is able to reduce or eliminate the impact of outliers. 

\end{abstract}

\section{Introduction}
Principal Component Analysis (PCA) \cite{pearson1901liii} is one of the key methods for dimensionality reduction and feature extraction in data analysis \cite{bishop2006pattern, jolliffe2011principal}, which plays an important role in machine learning \cite{murphy2012machine}, computer vison \cite{de2001robust}, genetics \cite{biffi2010principal} and so on. As high dimensional data is difficult to analyze effectively and brings a huge computing burden, PCA attempts to represent high-dimensional real data with principal components. 

There exist two common definitions of PCA, both of which are algebraic and lack probabilistic explanation for the observed data \cite{tipping1999probabilistic}. In order to overcome this shortcoming, Tipping and Bishop proposed a probabilistic model for PCA, called Probabilistic PCA (PPCA) \cite{tipping1999probabilistic}. PPCA has several advantages over traditional PCA, such as, it can deal with missing values in the observed data due to the related EM algorithm.

However, despite these merits, PCA and PPCA inherently have a drawback: they are not robust, i.e., they are forcefully affected by the outliers \cite{stanimirova2007dealing, serneels2008principal, zhang2014novel}. Due to the presence of outliers, the principal components obtained by classical PCA and PPCA are greatly deviated from the real direction. Consequently, we can not get the main information of the data exactly. Subsequent analysis based on these principal components can not obtain satisified results.

By imitating human or animal learning, Self-Paced Learning (SPL) usually begins with simple samples of learning tasks, then introduces complex examples into the training process step by step \cite{meng2015objective}. 
SPL has been applied in object detection \cite{zhang2017bridging, sangineto2019self}, matrix factorization \cite{zhao2015self}, mixture of regression \cite{han2017self}. It has been shown that SPL has the ability to reduce the effect of outliers \cite{zhang2018self}. 

In order to improve the robustness of the dimensionality reduction algorithms, we incorporate Self-Paced Learning mechanism \cite{kumar2010self} into PPCA, and propose a novel model called Self-Paced Probabilistic Principal Component Analysis (SP-PPCA). Based on PPCA, we design a new objective function by introducing additional parameters about self-paced learning. We use the alternative search strategy to learn the original parameters of PPCA and the additional introduced parameters. The proposed method attempts to learn from the clean training data gradually, and simultaneously prevent outliers from affecting the training process.

In summary, our main contributions in this paper are the following.
\begin{itemize}
    \item To effectively eliminate the impact of outliers, we introduce Self-Paced Learning mechanism into Probabilistic PCA, which is the earliest effort to use SPL for PPCA.
    \item A novel approach named SP-PPCA is constructed from a probabilistic perspective. We derive a corresponding optimization method based on the expectation maximization algorithm and the alternative search strategy.
    \item We conduct extensive experiments on both simulated and real data sets. The results show that the proposed method can obtain more accurate projection vectors from the contaminated data.
\end{itemize}

The remainder of this work is organized as follows. In Section \ref{sec2}, we briefly describe Probabilistic PCA. In Section \ref{sec3}, we propose the Self-Paced Probabilistic PCA. In Section \ref{sec4}, we evaluate our method by experiments on synthetic data and real data. In Section \ref{relatedwork}, we introduce some related work. Finally, we summarize the paper in Section \ref{sec5}.

\section{Probabilistic PCA}
\label{sec2}
PCA can be defined as the orthogonal projection of the data onto a lower dimensional linear space called the principal subspace, where the variance of the projected data is maximized \cite{hotelling1933analysis} \cite{bishop2006pattern}. An equivalent definition is the linear projection that minimizes the mean squared distance between the data points and their projections \cite{pearson1901liii}. 

The above two definitions are algebraic and lack probabilistic explanation for the observed data. PCA can also be expressed as the maximum likelihood solution of a probabilistic latent variable model \cite{bishop2006pattern}. The probabilistic version of PCA is known as Probabilistic PCA (PPCA) \cite{tipping1999probabilistic}.

Probabilistic PCA introduces a $M$-dimensional vector of latent variable $\mathbf{z}$ corresponding to the principal component subspace. The prior distribution of $\mathbf{z}$ is assumed to be:
$$
p(\mathbf{z}) = \mathcal{N} ( \mathbf{z} | \mathbf {0}_M , \mathbf{I}_M )
.
$$
The $D$-dimensional observed data vector $\mathbf{x}$ is formulated by a linear combination of the latent variable $\mathbf{z}$ plus noise $\bm{\epsilon}$:
\begin{equation}
      \mathbf{x} = \mathbf {W} \mathbf {z} + \bm{\mu} + \bm{\epsilon}
      ,
\label{x_linear}
\end{equation}
where $\mathbf{W}$ is a $D \times M$ matrix which relates the observation vector $\mathbf{x}$ and the latent variable $\mathbf{z}$; the $D$-dimensional vector $\bm{\mu}$ allows the model to have non-zero mean; $\bm{\epsilon}$ is a $D$-dimensional Gaussian-distributed variable, i.e., $\bm{\epsilon} \sim \mathcal{N}(\mathbf{0}_D, \sigma^{2} \mathbf{I}_D)$.

Hence, equation (\ref{x_linear}) induces the conditional distribution of the observed data vector $\mathbf{x}$:
$$
p (\mathbf{x} | \mathbf{z} ) = 
\mathcal{N} ( \mathbf{x} | \mathbf{W} \mathbf{z} + \bm{\mu} , \sigma^{2} \mathbf{I}_D )
.
$$
By integrating out the latent variables, the marginal distribution of the observed variable is obtained:
$$
p( \mathbf{x} ) 
= \int p( \mathbf{x} | \mathbf{z} ) p (\mathbf{z}) \mathrm{d} \mathbf{z}
= \mathcal{N} ( \mathbf{x} | \bm{\mu} , \mathbf{C} )
,
$$
where $\mathbf{C}$ is a $D \times D$ covariance matrix defined by
$
\mathbf{C} = \mathbf{W} \mathbf{W}^{\mathrm{T}} + \sigma^{2} \mathbf{I}_D.
$
To improve the efficiency of later calculations, we give the fact that
$
\mathbf{C}^{-1} = \sigma^{-2} \mathbf{I}_D - \sigma^{-2} \mathbf{W} \mathbf{M} ^ {-1} \mathbf{W} ^ {\mathrm{T}}
,
$
where the $M \times M$ matrix $\mathbf{M}$ is defined by
\begin{equation}
   \mathbf{M} = \mathbf{W} ^ {\mathrm{T}} \mathbf{W} + \sigma^{2} \mathbf{I}_M
.   
\label{define_M}
\end{equation}
So the cost of evaluating $\mathbf{C}^{-1}$ is reduced from $O(D^3)$ to $O(M^3)$ \cite{bishop2006pattern}. And the posterior distribution $p(\mathbf{z}|\mathbf{x})$ can be calculated using Bayes’ rule, which is a Gaussian distribution:
\begin{equation}
p ( \mathbf{z} | \mathbf{x} ) 
= 
\mathcal{N} ( \mathbf{z} | \mathbf{M}^{-1} \mathbf{W} ^ {\mathrm{T}} ( \mathbf{x} - \bm{\mu} ) , \sigma^{2} \mathbf{M}^{-1} )
.   
\label{distribution_zx}
\end{equation}

Finally, based on a data set $\mathbf{X} = \{\mathbf{x}_n, n = 1,2,\cdots,N\}$, where $\mathbf{x}_n$ is the vector presentation of observed data, the log likelihood function is given by
\begin{equation}
\begin{aligned}
&
\ln p(\mathbf{X} | \bm{\mu}, \mathbf{W}, \sigma^{2}) = 
\sum_{n=1}^{N} \ln p 
\left( 
\mathbf{x}_{n} | \bm{\mu} , \mathbf{W} ,  \sigma^{2} 
\right) \\
& = - \frac{ND}{2} \ln(2\pi) - \frac{N}{2} \ln|\mathbf{C}| - 
\frac{1}{2} \sum_{n=1}^{N} 
\left( \mathbf{x}_{n} - \bm{\mu} \right) ^ {\mathrm{T}} 
\mathbf{C}^{-1} 
\left( \mathbf{x}_{n} - \bm{\mu} \right) .
\end{aligned}
\label{llh}
\end{equation}
The model parameters $\{\bm{\mu}, \mathbf{W}, \sigma^2\}$ can be estimated using maximum likelihood. They have exact closed-form solutions, and can also be found through an EM algorithm\cite{tipping1999probabilistic, bishop2006pattern}.

\section{Self-Paced Probabilistic PCA}
\label{sec3}
In this section, we explain the proposed method and the corresponding optimization algorithm in detail.

\subsection{Object Function}
We incorporate the Self-Paced Learning mechanism to PPCA, and propose Self-Paced Probabilistic Principal Component Analysis (SP-PPCA). The objective function of SP-PPCA is defined based on the above optimization problem (\ref{llh}) by introducing binary variables $v_n \in \{0, 1\}$ ($n = 1,2,\cdots,N$) and adding a sparse regularizer of $v_n$, which is given by:
\begin{equation}
\mathcal{L}(\mathbf{v} , \bm {\mu} , \mathbf{W} , \sigma^2)
= 
\sum _ { n = 1 } ^ { N }
v_n
l_n
-
\beta \sum_{n=1}^{N} v_n
,
\label{sp_objective}
\end{equation}
where $l_n = - \ln p (\mathbf{x}_n | \bm{\mu}, \mathbf{W}, \sigma^2)$, $\mathbf{v} = (v_1, v_2, \cdots, v_N)^{\mathrm{T}}$, $\beta$ is a hyper-parameter. The introduced binary variable $v_n$ indicates whether the $n^{th}$ sample $\mathbf{x}_n$ is an outlier (if $v_n=0$) or a clean sample (if $v_n=1$). Note that, like \{$\bm{\mu} , \mathbf{W} , \sigma^2$\},  $\mathbf{v}$ also needs to be estimated from data. In SP-PPCA, the goal is to \textbf{minimize} equation (\ref{sp_objective}). In the next section \ref{optimization_v}, we will describe how to learn the parameters \{$\bm{\mu} , \mathbf{W} , \sigma^2$\} and $\mathbf{v}$ iteratively.

\subsection{Optimization}
Based on the Self-Paced Learning strategy, we solve the minimization problem (\ref{sp_objective}) by beginning with simple data points, then introducing complex examples into training gradually, meanwhile filtering out outliers during the training process. 

For each fixed $\beta$, we use the alternative search strategy to obtain approximate solution of the above problem efficiently. Specifically, given parameters \{$\bm{\mu}$, $\mathbf{W}$, $\sigma^2$\}, we estimate $\mathbf{v}$; and for a fixed $\mathbf{v}$, we update parameters $\{\bm{\mu}, \mathbf{W}, \sigma^2\}$. A brief description of the algorithm is presented in Algorithm \ref{algorithm_all}. In the following, we describe each component in detail.

\subsubsection{\uppercase\expandafter{\romannumeral1}. Optimization of $\mathbf{v}$.}
\label{optimization_v}
Fixing parameters \{$\bm{\mu}$, $\mathbf{W}$, $\sigma^2$\}, we estimate $\mathbf{v}$ by solving the following problem:
$$
\min \limits_{\mathbf{v} \in \{0,1\}} 
\mathcal{L} ( \mathbf{v} ; \bm{\mu} , \mathbf{W} , \sigma^{2} )
,
$$
which is equivalent to
$$
\min \limits_{\mathbf{v} \in \{0,1\}}
\sum_{n=1}^N v_n (l_n - \beta)
.
$$
Thus the solution of $\mathbf{v}$ can be easily obtained by:
\begin{equation}
v_{n} = 
\left
\{ 
\begin{array}{lll} 
{0} & , &{ l_n > \beta }, \\
& &\\
{1} & , &{ l_n \leq \beta. } 
\end{array} 
\right.
\label{solution_v}
\end{equation}
Then, the following two important questions can be answered.
\begin{itemize}
      \item Why can $\mathbf{v}$ be viewed as the outlier indicator?
      \item What role does the hyper-parameter $\beta$ play?
\end{itemize}

On the basis of equation (\ref{solution_v}), we find that $\mathbf{v}$ can be estimated simply based on $\beta$ as a threshold. When $l_n$ is larger than $\beta$, $v_n$ is set to 0. Since outliers are usually located far from the data center, they generally have lower likelihood according to equation (\ref{llh}). Thus, the training sample $\mathbf{x}_n$ is more likely to be an outlier if the sample with larger $l_n$, i.e., lower likelihood. Therefore, it is reasonable to use $\mathbf{v}$ to indicate outliers in the training procedure.

It can be seen that $\beta$ is a very crucial hyper-parameter. If $\beta$ is small, the optimization problem prefers to only consider relatively ``cleaner" samples with high likelihood; on the contrary, most of samples (maybe also include outliers) are introduced if $\beta$ is very large. Thus, we use an adaptive strategy to tune $\beta$, as shown in Algorithm \ref{algorithm_all}. We increase the value of $\beta$ by a factor $\eta$ iteratively until the objective function value $\mathcal{L}(\mathbf{v} , \bm {\mu} , \mathbf{W} , \sigma^2)$ converges. 
This strategy can collect clean data for training, at the same time, prevent outliers from skewing the results.

\subsubsection{\uppercase\expandafter{\romannumeral2}. Optimization of $\{\bm{\mu}, \mathbf{W}, \sigma^2\}$.}
Given fixed $\mathbf{v}$, we estimate $\{\bm{\mu}, \mathbf{W}, \sigma^2\}$ by
$$
\min \limits_{ \bm{\mu}, \mathbf{W},\sigma^{2}} 
\mathcal{L} ( \bm{\mu} , \mathbf{W} , \sigma^2 ; \mathbf{v} ) 
,
$$
which is equivalent to
$$
\max \limits_{\bm{\mu}, \mathbf{W}, \sigma^{2}} 
\sum_{n=1}^N
v_n \ln p (\mathbf{x}_n | \bm{\mu}, \mathbf{W}, \sigma^2) 
.
$$

It can be seen the above problem is similar to equation (\ref{llh}) in PPCA. By setting the derivative of the above equation with respect to $\bm{\mu}$ equal to zero, we get:
\begin{equation}
\bm{\mu}_{\text{new}} = \frac{\sum_{n=1}^N v_n \mathbf{x}_n}{\sum_{n=1}^N v_n}
,
\label{mu}
\end{equation}
and then substitute $\bm{\mu}$ with $\bm{\mu}_{\text{new}}$. Next, we use the EM algorithm to maximize the problem with respect to $\mathbf{W}$, $\sigma^2$. The complete-data log likelihood function of this problem is given by:
$$
\ln 
p(\mathbf{X}, \mathbf{Z} | \bm{\mu}, \mathbf{W}, \sigma^2) =
\sum_{n=1}^N  v_n
\Big\{
\ln p(\mathbf{x}_n | \mathbf{z}_n) + \ln p (\mathbf{z}_n)
\Big\}
,
$$
where $\mathbf{z}_n^{\mathrm{T}}$ is the $n^{th}$ row of the matrix $\mathbf{Z}$. Then we calculate the expectation with respect to the posterior distribution of the latent variables as follows,
$$
\begin{aligned}
\mathbb{E} 
[\ln p ( \mathbf{X} , \mathbf{Z} | \bm{\mu} , \mathbf{W} , \sigma^2 ) ] 
= &
- \sum_{n=1}^{N} v_n 
\bigg\{
\frac{D}{2} \ln 
\left( 2 \pi \sigma^2 \right) 
+ 
\frac{1}{2} \operatorname{Tr} 
\left( 
\mathbb{E} 
\left[ \mathbf{z}_{n} \mathbf{z}_{n}^{\mathrm{T}} \right] 
\right)\\ 
&+ \frac{1}{2 \sigma^2} 
\left\| \mathbf{x}_{n} - \bm{\mu} \right\| ^ 2
- \frac{1}{\sigma^2} 
\mathbb{E} 
\left[ \mathbf{z}_{n} \right] ^ {\mathrm{T}} 
\mathbf{W}^{\mathrm{T}} 
\left( \mathbf{x}_{n} - \bm{\mu} \right) \\
&+ \frac{1}{2\sigma^2} \operatorname{Tr} 
\left( 
\mathbb{E} \left[ \mathbf{z}_{n} \mathbf{z}_{n}^{\mathrm{T}} \right] 
\mathbf{W}^{\mathrm{T}} \mathbf{W} 
\right) 
+ \frac{M}{2} \ln (2\pi)
\bigg\}
.
\end{aligned}
$$

From the above derivation, in the \underline{\textbf{E step}} of the EM algorithm, we compute
\begin{equation}
\begin{aligned} 
\mathbb{E} \left[ \mathbf{z}_{n} \right] & = 
\mathbf{M}^{-1} \mathbf{W}^{\mathrm{T}} 
\left( 
\mathbf{x}_{n} - \bm{\mu}_{\text{new}}
\right) 
,
\\ 
\mathbb{E} 
\left[ \mathbf{z}_{n} \mathbf{z}_{n}^{\mathrm{T}} \right] 
& = 
\sigma^2 \mathbf{M}^{-1} + 
\mathbb{E} \left[ \mathbf{z}_{n} \right] 
\mathbb{E} \left[ \mathbf{z}_{n} \right] ^ {\mathrm{T}} 
,
\end{aligned}  
\label{Estep} 
\end{equation}
where $\mathbf{M}$ is defined by equation (\ref{define_M}). The above equations can be obtained easily using the posterior distribution (\ref{distribution_zx}) of latent variables.
Then, in the \underline{\textbf{M step}}, by setting the derivatives with respect to $\mathbf{W}$ and $\sigma^2$ to zero respectively, we obtain the M-step re-estimation solutions:
\begin{equation}
\mathbf{W}_{\text{new}} 
= 
\left[ 
\sum_{n=1}^{N}  v_n
\left( 
\mathbf{x}_{n}- \bm{\mu}_{\text{new}} 
\right) 
\mathbb{E} \left[ \mathbf{z}_{n} \right] ^ {\mathrm{T}} \right] 
\left[ 
\sum_{n=1} ^ {N} v_n
\mathbb{E} 
\left[ \mathbf{z}_{n} \mathbf{z}_{n} ^ {\mathrm{T}} \right] 
\right] ^ {-1} 
,  
\label{Mstep_W}
\end{equation}
\begin{equation}
\begin{aligned} 
\sigma_{\text{new}}^{2} 
=  \frac{1}{D \sum_{n=1}^N v_n} 
\sum_{n=1}^{N}  v_n
\Big\{ 
&
\left\| \mathbf {x}_{n} - \bm{\mu}_{\text{new}} \right\| ^ {2} 
+ \operatorname{Tr} 
\left( \mathbb{E} 
\left[ \mathbf{z}_{n} \mathbf{z}_{n} ^ {\mathrm{T}} \right] 
\mathbf{W}_{\text{new}} ^ {\mathrm{T}} 
\mathbf{W}_{\text{new}} 
\right) 
\\
& - 2 \mathbb{E} \left[ \mathbf{z}_{n} \right] ^ {\mathrm{T}} 
\mathbf{W}_{\text{new}} ^ {\mathrm{T}} 
\left( \mathbf{x}_{n} - \bm{\mu}_{\text{new}} \right) 
\Big\}.
\end{aligned}
\label{Mstep_sigma}
\end{equation}

From equations (\ref{mu}), (\ref{Mstep_W}) and (\ref{Mstep_sigma}), only data points indicated as ``clean'' affect the values of the parameters $\{\bm{\mu}, \mathbf{W}, \sigma^2\}$. In other words, the projection vectors obtained by SP-PPCA are rarely influenced by outliers.

\subsubsection{\uppercase\expandafter{\romannumeral3}. Summary.}
The overall algorithm is shown in Algorithm \ref{algorithm_all}. At each iteration, we increase the value of $\beta$ through a factor $\eta$ until the objective function (\ref{sp_objective}) converges. For a fixed $\beta$, we update the binary outlier indicator $\mathbf{v}$ and the original model parameters $\{\bm{\mu}, \mathbf{W}, \sigma^2\}$ iteratively. Thus, the proposed method SP-PPCA looks for optimal projection vectors and filters out outliers iteratively. Note that we find that changing $\beta$ immediately after updating parameters $\{\bm{\mu}, \mathbf{W}, \sigma^2\}$ and $\mathbf{v}$, in other words, keeping only the outer loop, can reduce the computational costs without affecting the performance significantly in our later experiments.
\begin{algorithm}[H] 
\caption{Self-Paced Probabilistic PCA}  
\begin{algorithmic}[1]
\Require Dataset $\mathbf{X}=\{ \mathbf{x_n}, n=1,2,\cdots,N \}$, learning rate $\eta > 1$.
\Ensure $\bm{\mu}$, $\mathbf{W}$, $\sigma^2$.
\State Initialize $\bm{\mu}$, $\mathbf{W}$, $\sigma^2$.\\
\Comment{By the mean of samples, a random matrix, and scalar 1, respectively.}
\State Initialize $\beta$ to the median of $l_n, n = 1,2,\cdots,N$. \\
\begin{flushright}
      \Comment{$l_n$ for initialization is obtained by running only\\ one iteration of the original PPCA algorithm.}
\end{flushright}
\Repeat
	\Repeat
      \State Update $\mathbf{v}$ by equation (\ref{solution_v}).
      \State Update $\bm{\mu}$, $\mathbf{W}$, $\sigma^2$ by equations (\ref{mu}), (\ref{Estep}), (\ref{Mstep_W}), (\ref{Mstep_sigma}).
	\Until{convergence}
	\State $\beta \leftarrow \eta \beta$.
\Until{convergence}
\end{algorithmic}
\label{algorithm_all}
\end{algorithm}

Once we obtain the convergence results $\{\bm{\mu}_{*}, \mathbf{W}_{*}, {\sigma^2}_{*}\}$, a $D$-dimensional point $\mathbf{x}$ in the data space can be represented by the corresponding posterior mean and covariance in the latent space \cite{bishop2006pattern} according to equation (\ref{distribution_zx}). The mean is obtained by:
$$
\mathbb{E}[\mathbf{z} | \mathbf{x}] = \mathbf{M}_{*}^{-1} \mathbf{W}_{*}^{\mathrm{T}} (\mathbf{x} - \bm{\mu}_{*})
,
$$
where $\mathbf{M}_{*} = \mathbf{W}_{*}^{\mathrm{T}} \mathbf{W}_{*} + \sigma^2_{*} \mathbf{I}_{M}$.
We also can reconstruct the original data point $\mathbf{x}$ by $\mathbf{\hat{x}}$:
$$
\mathbf{\hat{x}} = \mathbf{W} \mathbb{E}[\mathbf{z} | \mathbf{x}] + \bm{\mu}_{*}.
$$

\section{Experiments}
\label{sec4}
The goal of our work is to get the correct principal components that are not influenced much by outliers. We follow the evaluation method in \cite{minnehan2019grassmann, ju2015image}.
Formally, we have a contaminated train dataset $\mathbf{X}_{\text{train}}$ and a clean test dataset $\mathbf{X}_{\text{test}}$. We perform dimensionality reduction on the contaminated data $\mathbf{X}_{\text{train}}$, and obtain projection vectors. Then we calculate the reconstruction error on the test data by 
\begin{equation}
Error = \frac{ ||\mathbf{X}_{\text{test}} - \mathbf{\widehat{X}}_{\text{test}}||_F }{||\mathbf{X}_{\text{test}}||_F}
,
\label{simu_error}
\end{equation}
where, $\mathbf{\widehat{X}}_{\text{test}}$ is the recovered data for $\mathbf{X}_{\text{test}}$ based on the obtained projection vectors, $||\cdot||_F$ is the Frobenius norm. A small reconstruction error means that the projection vectors obtained by dimensionality reduction methods contain more favorable information about true data and less negative information about outliers.

The proposed algorithm SP-PPCA is tested on both simulated data and real data with classical PCA, PPCA \cite{tipping1999probabilistic}, PCP \footnote{The PCP algorithm decomposes the observed data into a low rank matrix and a sparse matrix. We treat the sparse matrix as noise part, and perform standard PCA on the low rank matrix. Note that PCP is not specifically designed for data with outliers.} \cite{xiao2017online, candes2011robust}, RAPCA \cite{hubert2002fast}, ROBPCA \cite{hubert2005robpca} and $L_1$-norm PCA \cite{markopoulos2017efficient} as baseline methods \footnote{We implemented PCA using the module in the Sklearn library \cite{scikit-learn}. PPCA was implemented based on code in \url{http://prml.github.io/}. The code for PCP was obtained from \url{https://github.com/wxiao0421/onlineRPCA}. The codes of RAPCA and ROBPCA were obtained from \url{https://wis.kuleuven.be/stat/robust/LIBRA/LIBRA-home}. The code of $L_1$-norm PCA from \url{https://ww2.mathworks.cn/matlabcentral/fileexchange/64855-l1-pca-toolbox}. }. 
The code of our method is publicly available at \url{https://github.com/rumusan/SP-PPCA}. 

We began our empirical evaluations by exploring the performance over two synthetic experiments. In section \ref{twodim}, we tested on two-dimensional data firstly so that we could visualize the projection vectors. In section \ref{lowrank}, to further demonstrate the robustness of SP-PPCA, the experiments on high dimensional artifical data were analyzed. Then, in section \ref{ISOLETsec} and \ref{yalesec}, we compared the performance of SP-PPCA with other methods on two real datasets respectively: the ISOLET dataset and the Yale face dataset. In our experiments, the methods were based on default parameters and all experiments were averaged over 5 random trials.

\subsection{Experiments on two-dimensional data}
\label{twodim}
In this section, we compared our method with six different algorithms on two-dimensional data.
In the original data set $\{x_n, y_n\}$, $\{x_n\}$ were sampled from a uniform distribution from 0 to 150, and $\{y_n\}$ were draw from $y_n = 0.8 \times x_n + 5 + \epsilon$, where the noise $\epsilon$ was generated from normal distribution $\mathcal{N}(0,3)$. Then we added some outliers to the original data as contaminated data. We reduced the clean original data and the dirty data to one dimension by different approaches.
\begin{figure}[tbp]
\centering
\includegraphics[width=0.58\textwidth]{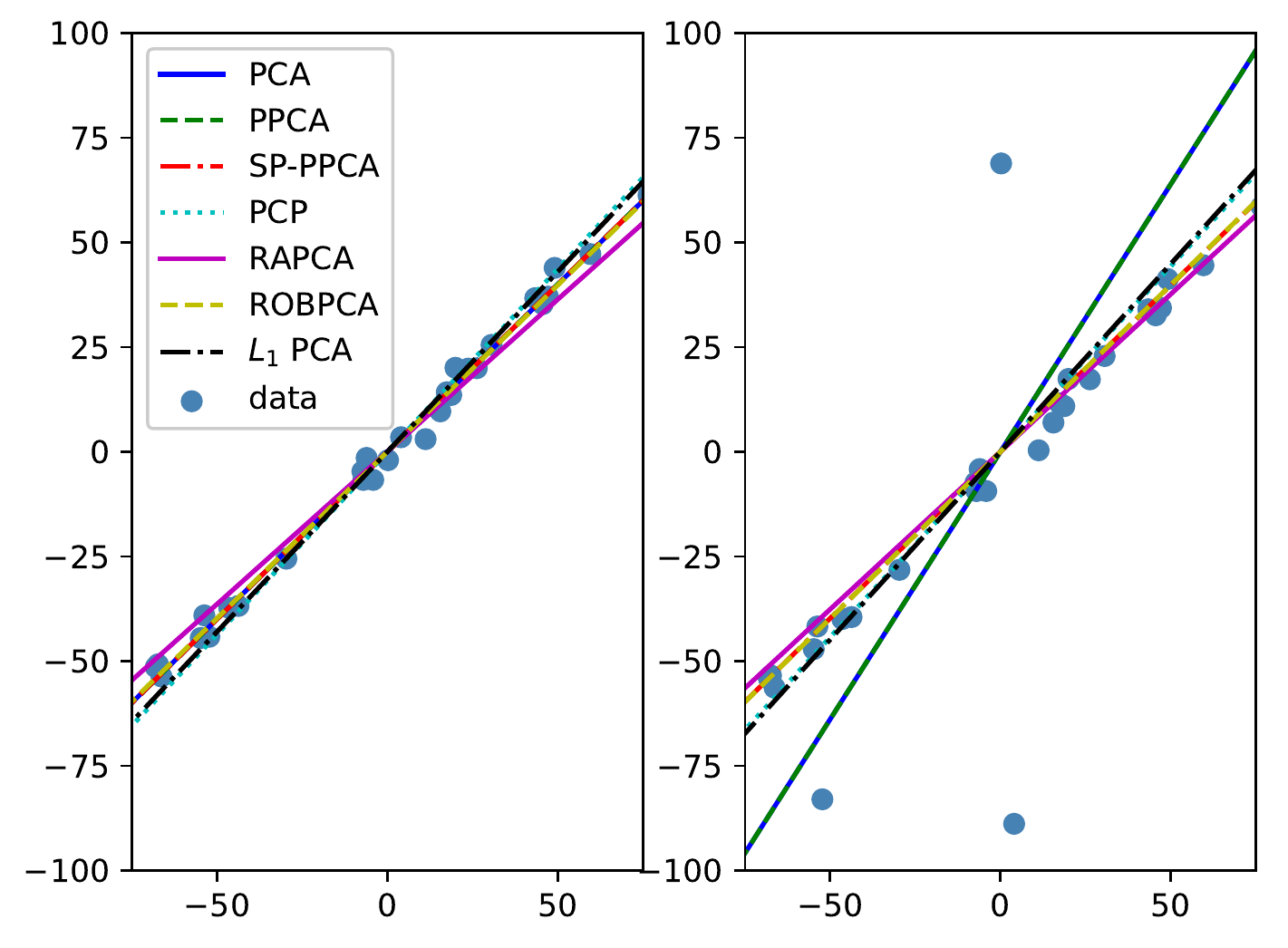}
\caption{The projection vectors obtained by different algorithms from the clean data (left) and the dirty data (right). The plotted data is centered. 
}
\label{fig_2dim}
\end{figure}

Figure \ref{fig_2dim} plots projection vectors obtained by these algorithms on both the clean data and the dirty data. It can be seen from this figure that the performance of these methods are similar on the clean data. However when it comes to the dirty data, PCA and PPCA are seriously affected by outliers. The outliers cause the projection vectors of PCA and PPCA to deviate from the correct direction. The results of this experiments show that the proposed method is superior to the traditional PCA and PPCA, and is comparable to PCP, RAPCA, ROBPCA and $L_1$-norm PCA.

\subsection{Experiments on low-rank matrices}
\label{lowrank}
We built a low-rank matrix as $\mathbf{X}_{raw} = \mathbf{U} \mathbf{V}^{\mathrm{T}} + \mathbf{E}$, where $\mathbf{U}$ (or $\mathbf{V}$) were taken as $N$ (or $D$) cases from a multivariate normal distribution $\mathcal{N}(\mathbf{0}, \mathbf{I})$; $\mathbf{E}$ is a $N \times D$ noise matrix, each row of which were generated from multivariate normal distribution $\mathcal{N}(\mathbf{0}_D, \mathbf{I}_D)$. Besides, to keep the level of noise low, the values of $\mathbf{E}$ were multiplied by 0.01. The dataset was divided into training set (70\%) and test set (30\%). Then the contaminated training data set $\mathbf{X}_{\text{train}}$ were constructed by replacing some normal samples in the original train set with outliers, which came from multivariate distribution $\mathcal{N}(\mathbf{1}_D, 5\times \mathbf{I}_D)$. We constructed a set of data with different sizes and various percentage of outliers. Then we run SP-PPCA and other algorithms on the occluded train data, and calculated reprojection errors on the test data.

\begin{figure}[htb]
\centering
\subfloat[]{\includegraphics[width=0.44\textwidth]{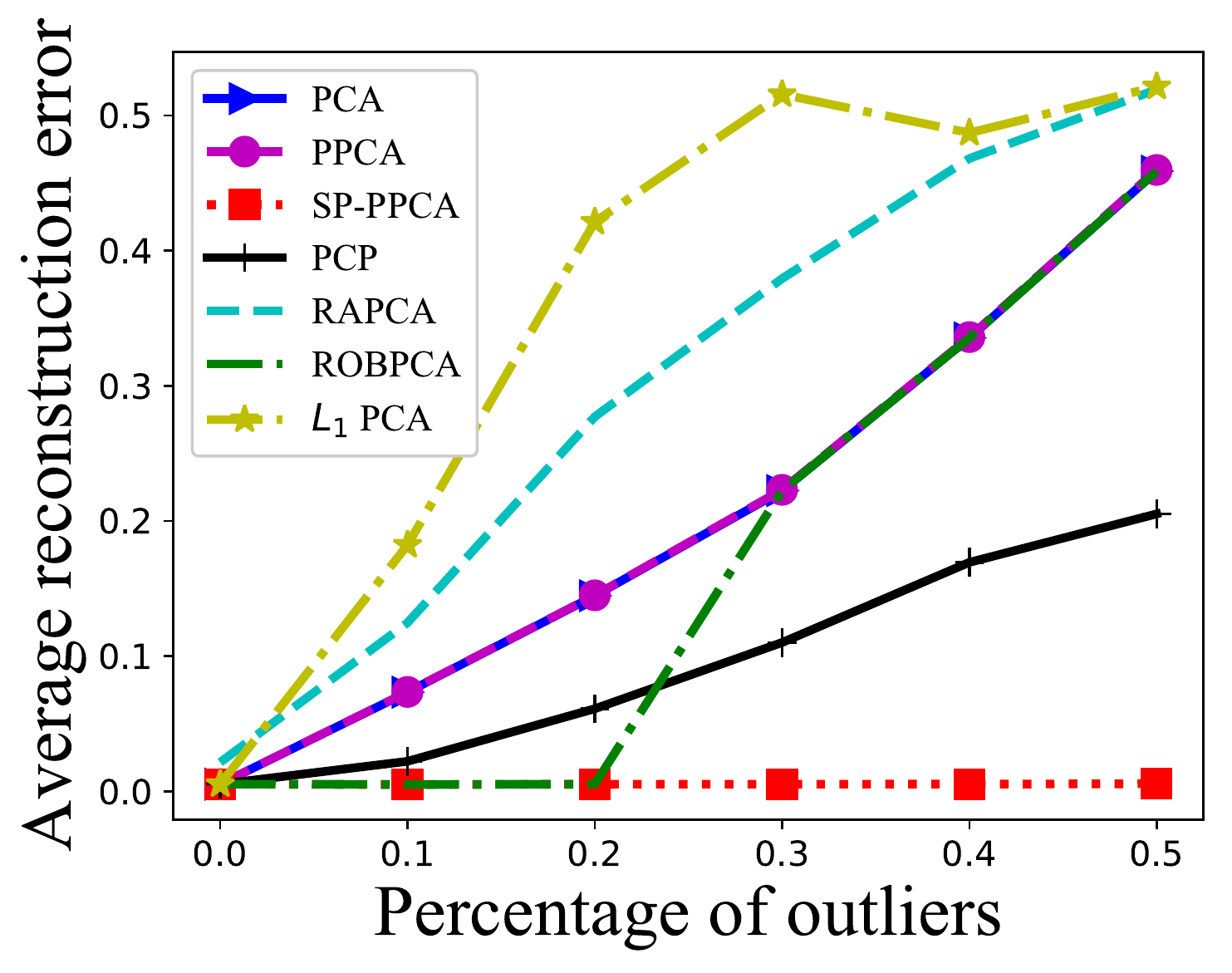}}
\subfloat[]{\includegraphics[width=0.44\textwidth]{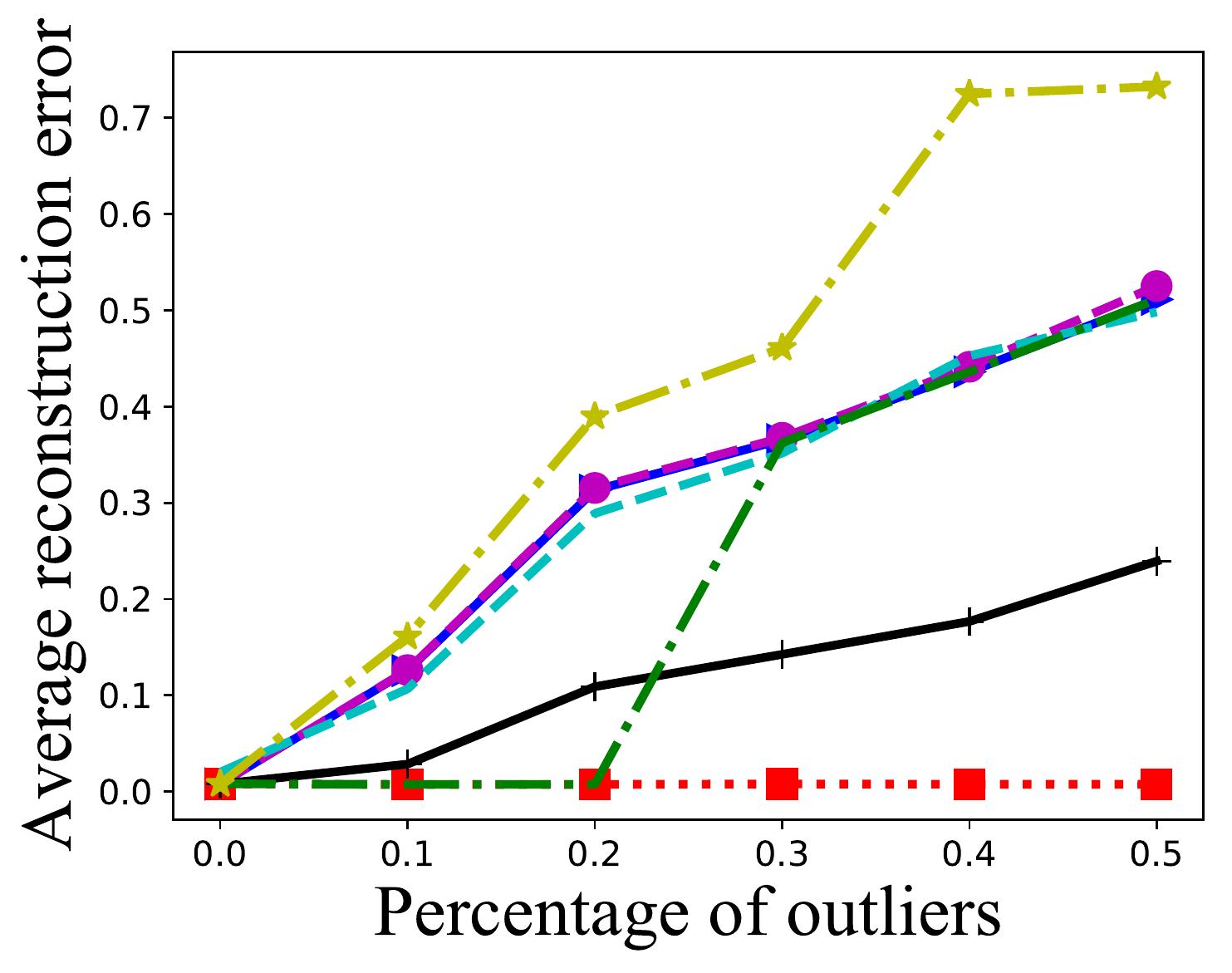}}\\
\subfloat[]{\includegraphics[width=0.44\textwidth]{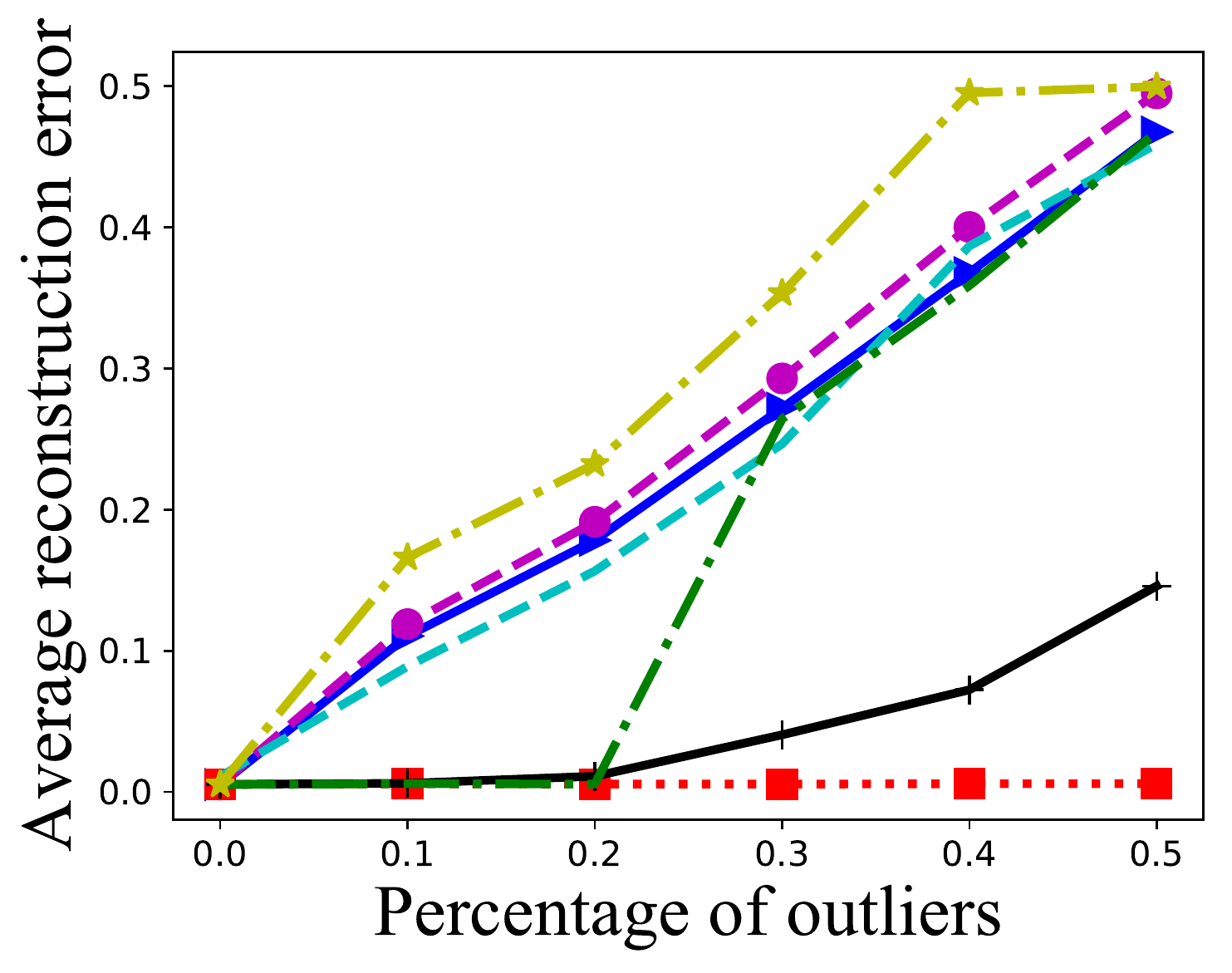}}
\subfloat[]{\includegraphics[width=0.44\textwidth]{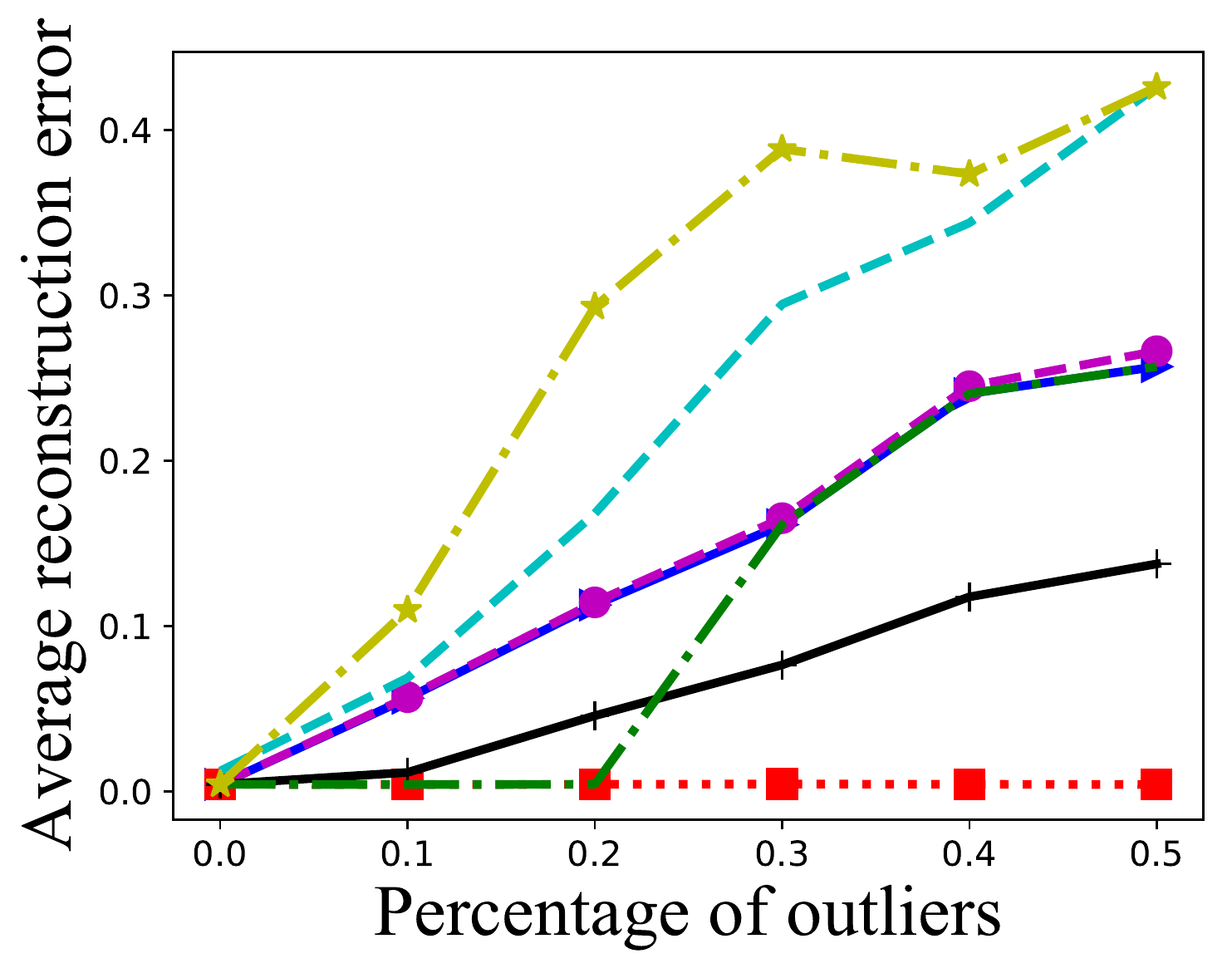}}
\caption{Average reconstruction errors for simulation problems with varying sizes of the low-rank matrices and different number of projection vectors. 
(a) $\mathbf{X}_{raw} \in \mathbb{R}^{100\times 200}$, 4 projection vectors;
(b) $\mathbf{X}_{raw} \in \mathbb{R}^{50\times 50}$, 2 projection vectors;
(c) $\mathbf{X}_{raw} \in \mathbb{R}^{100\times 20}$, 3 projection vectors;
(d) $\mathbf{X}_{raw} \in \mathbb{R}^{200\times 80}$, 5 projection vectors.
}
\label{fig_simulation_results}
\end{figure}

Figure \ref{fig_simulation_results} shows the average reconstruction errors by varying the sizes of the low-rank matrices and the number of projection vectors. From this figure we can see that the different methods perform similarly on the clean data. Nevertheless, PCA, PPCA, PCP, RAPCA, ROBPCA and $L_1$-norm PCA are greatly impacted by the outlying data points in dirty data. The proposed method can effectively withstand the influence of outliers and maintain low reconstruction errors.

\subsection{Experiments on ISOLET dataset}
\label{ISOLETsec}
Next, in section \ref{ISOLETsec} and \ref{yalesec}, we compared the performance of SP-PPCA with other methods \footnote{We did not use the $L_1$-norm PCA method for these experiments, because it is too time-consuming.} on real datasets.

In the third attempt, we conducted experiments on the ISOLET \footnote{\url{http://archive.ics.uci.edu/ml/datasets/ISOLET}} data set from UCI repository\cite{Dua:2019}. ISOLET is a dataset for spoken letter recognition with 7797 instances, and 617 attributes per sample. We randomly selected 900 samples as the training set and 600 samples as the test set. Then, we ``contaminate" our clean train data by some outliers. The attribute values of outliers were randomly sampled from the uniform distribution between the minimum and maximum value of the train data. Then dimensionality reduction algorithms were used to project the data to a low dimensional space. Finally, the reconstruction errors can be calculated by (\ref{simu_error}).

\begin{figure}[htb]
      \centering
      \subfloat[]{\includegraphics[width=0.44\textwidth]{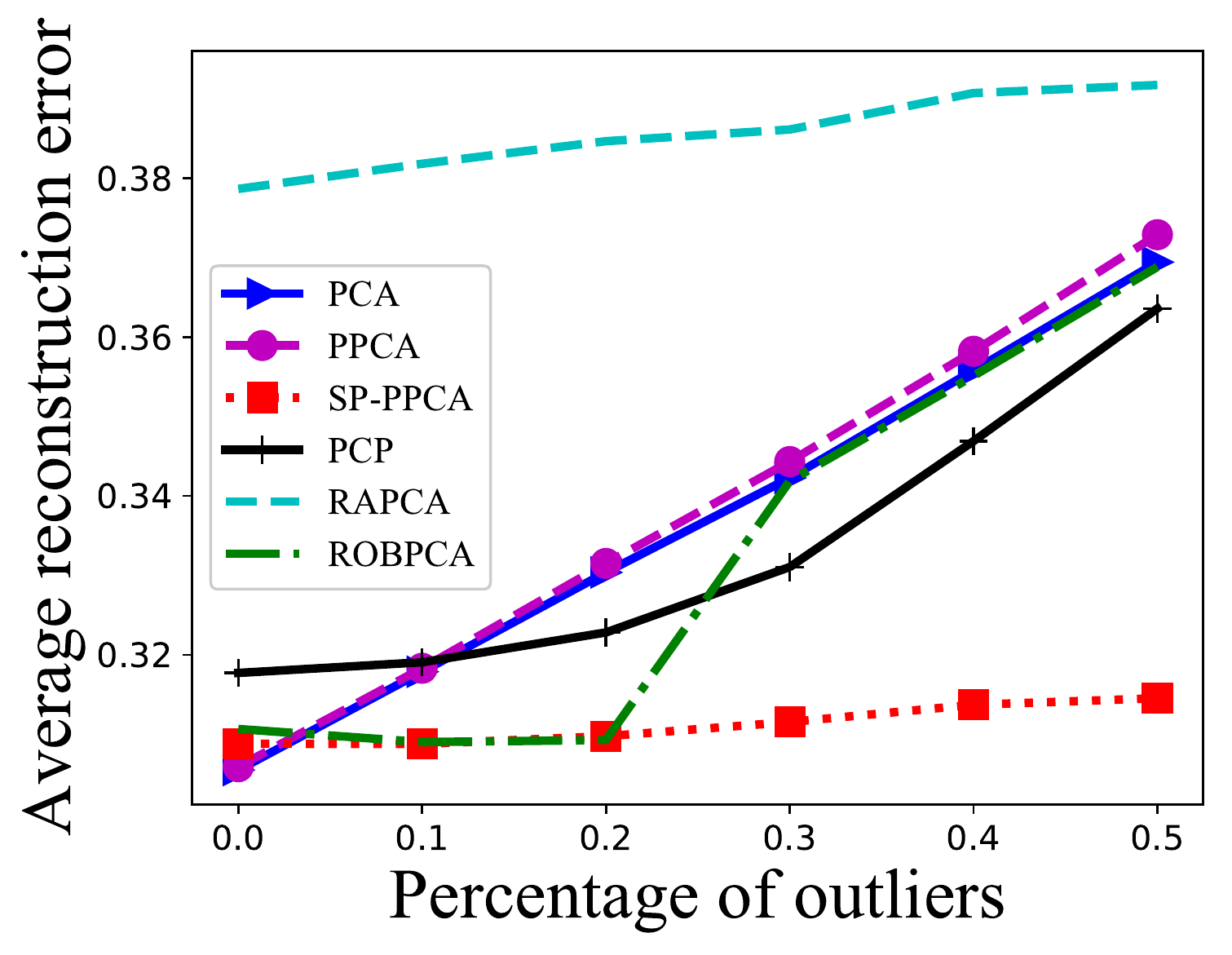}}
      \subfloat[]{\includegraphics[width=0.44\textwidth]{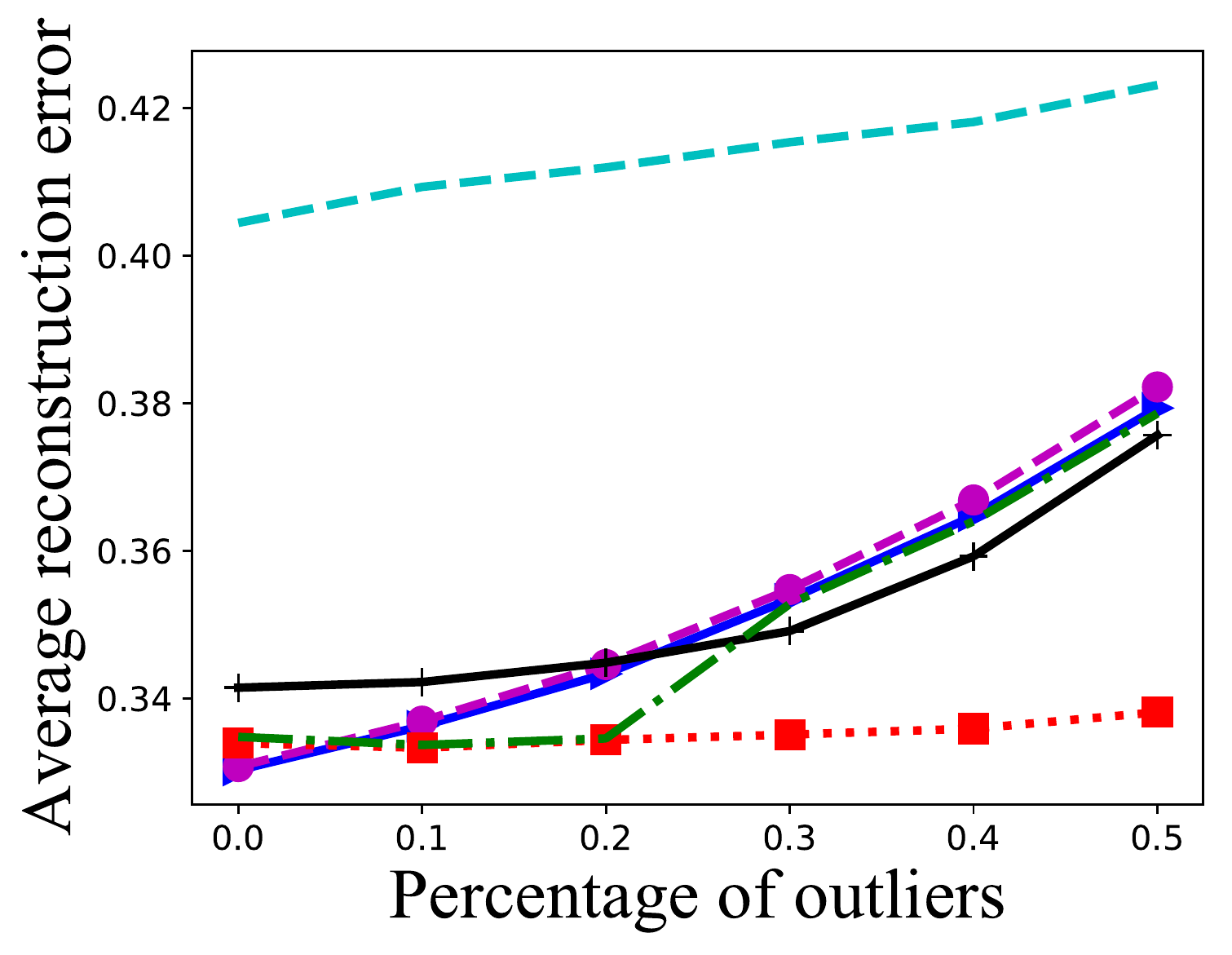}}\\
      \subfloat[]{\includegraphics[width=0.44\textwidth]{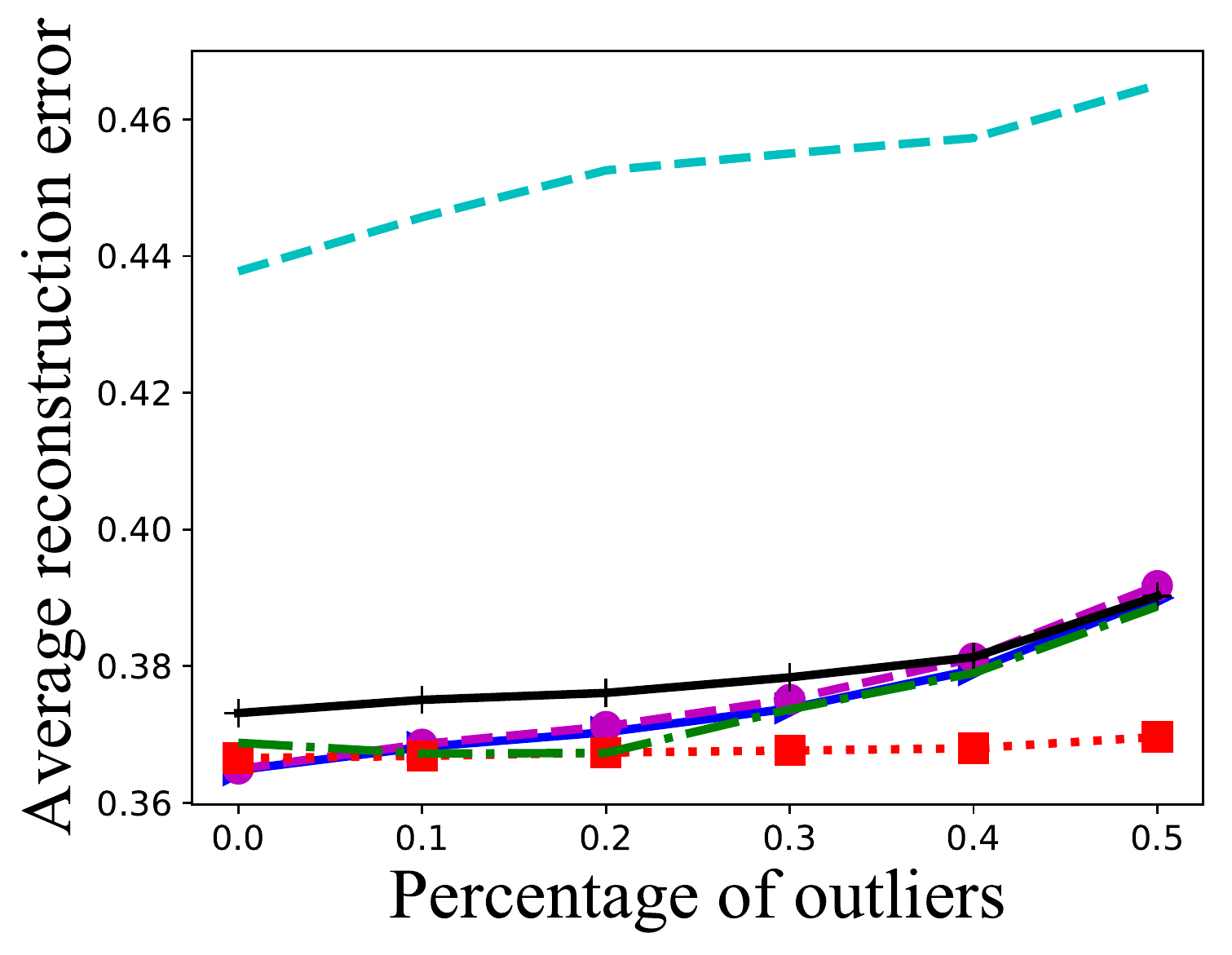}}
      \subfloat[]{\includegraphics[width=0.44\textwidth]{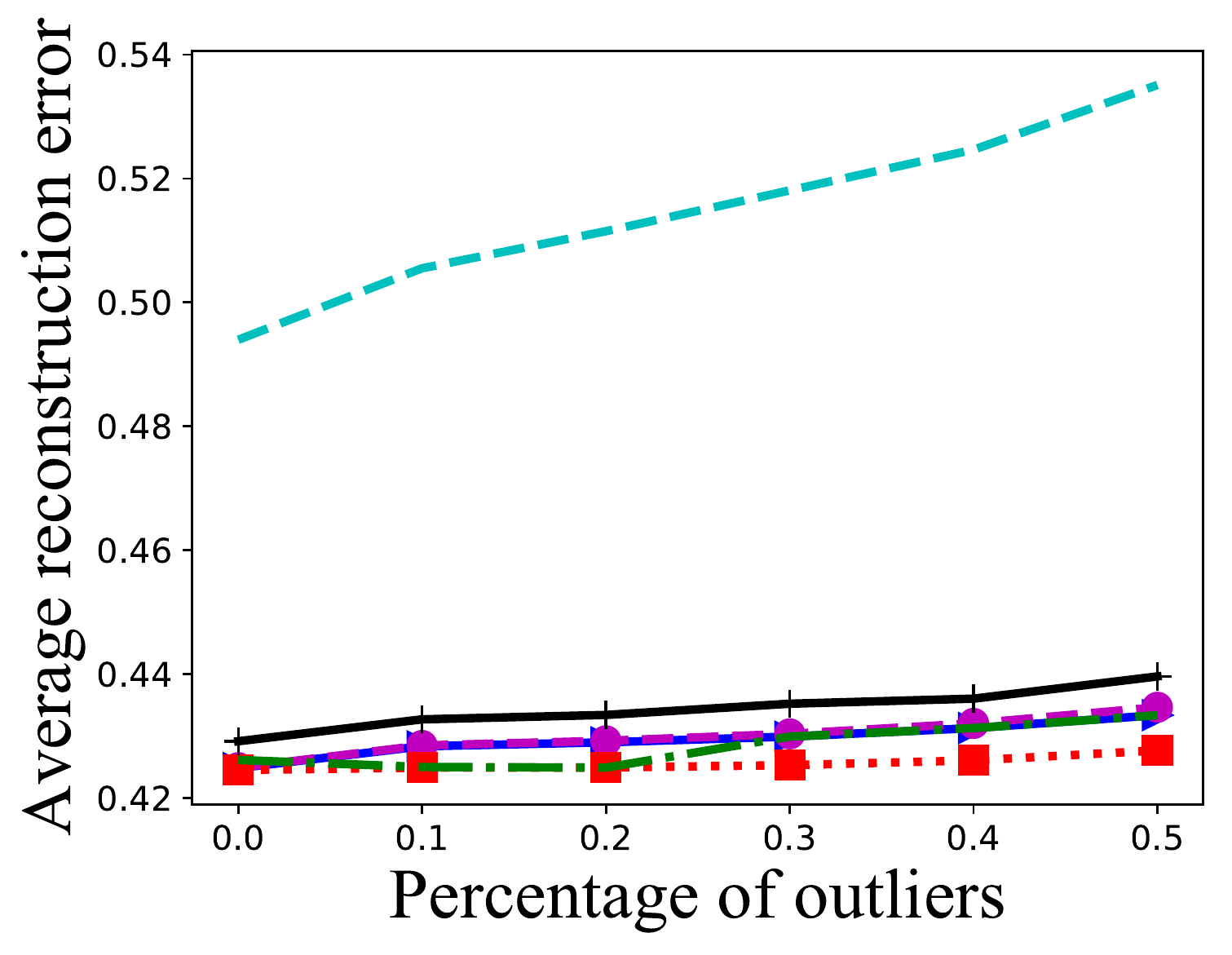}}
      \caption{Average reconstruction errors for ISOLET with different number of principal components.
      (a) 40 principal components;
      (b) 30 principal components;
      (c) 20 principal components;
      (d) 10 principal components.
      }
      \label{fig_isolet_results}
\end{figure}

Figure \ref{fig_isolet_results} presents the average reconstruction errors with different number of principal components. The performances of most methods deteriorate rapidly as the number of outliers increases, which is similar to the results of our simulation experiments in section \ref{lowrank}. SP-PPCA tends to remove outliers from the dirty dataset, thus, as shown in Figure \ref{fig_isolet_results}, the proposed method can get good results on the contaminated data as well as on the raw data.

\subsection{Experiments on Yale Face dataset}
\label{yalesec}
\begin{figure}[htbp]
\centering
\includegraphics[width=0.77\textwidth]{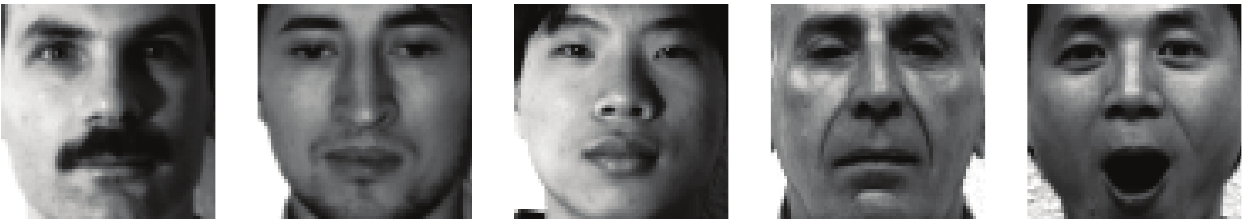} \\ 
\includegraphics[width=0.77\textwidth]{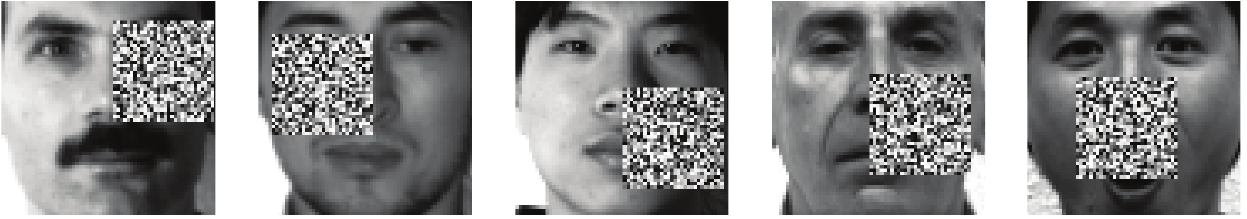}
\caption{Facial images in Yale Face dataset (the first row) and their corresponding noisy images (the second row).}
\label{fig_face_outlier}
\end{figure}

We further used the well-known Yale Face dataset \footnote{\url{http://cvc.cs.yale.edu/cvc/projects/yalefaces/yalefaces.html}} \cite{belhumeur1997eigenfaces} for our evaluation. The Yale Face dataset contains 165 grayscale images of 15 individuals. There are 11 images per subject with different illumination and expression. Each facial image is in 256 gray scales per pixel. All the images were scaled to $64\times 64$ pixels in experiments. The first row of Figure \ref{fig_face_outlier} shows some facial images in the Yale Face dataset. In experiments, we selected 9 images of each subject as the training data, and the rest 30 images make up the test set. 
We adopted the approaches in \cite{ju2015image, he2011robust} to produce noise on a small part of the original training images as outlier samples. 
More specifically, we randomly selected images from the original train set, then the selected images were occluded with rectangular blocks in random position consisting of random black and white dots. The number of outlier samples in the training set is 15 or 30, and the size of the noise blocks is $30\times 30$ or $45\times 45$ in our experiments. 
The corresponding noisy images of the original facial images are shown in the second row of Figure \ref{fig_face_outlier}. We executed dimensionality reduction algorithms on the data with outliers, and calculated reconstruction errors on the clean test data set.

\begin{figure}[htbp]
      \centering
      \includegraphics[width=0.13\textwidth]{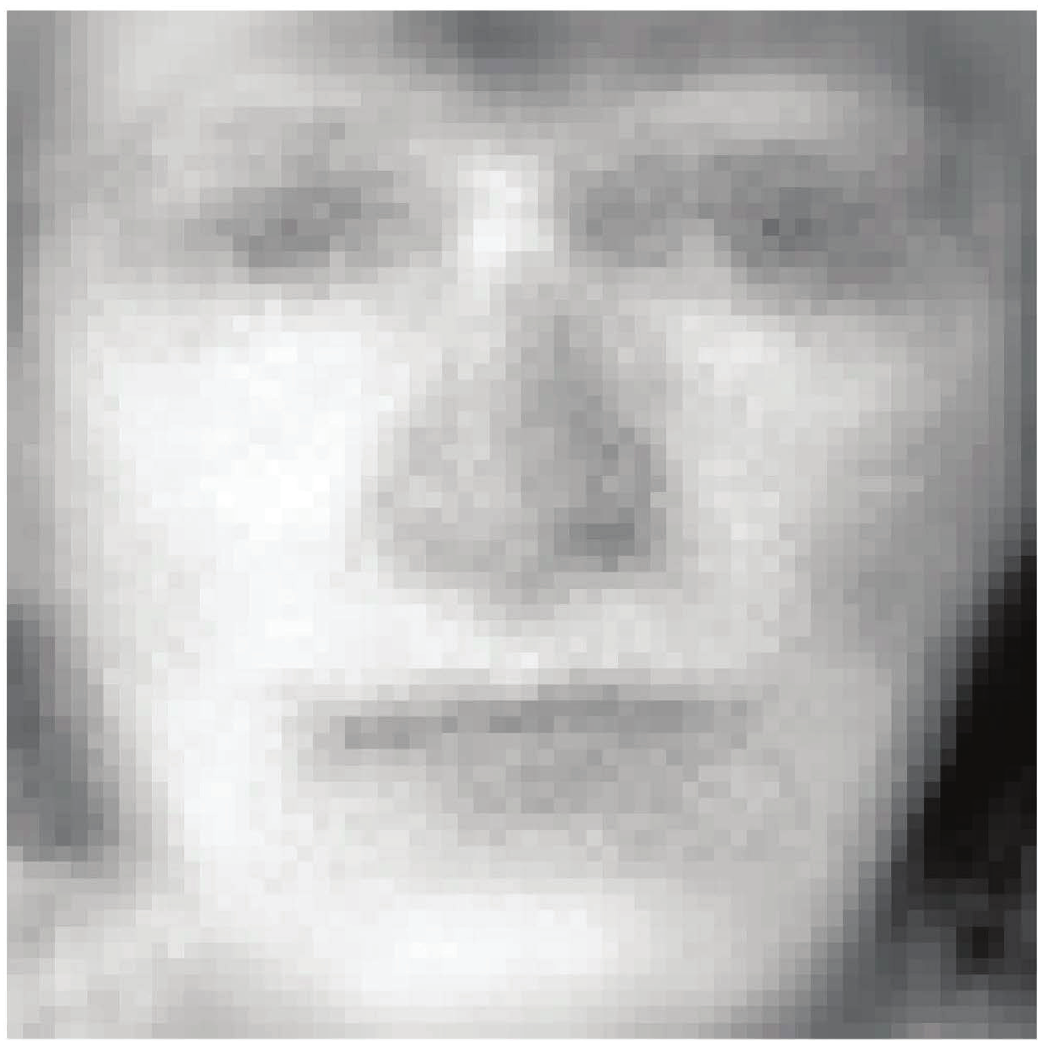}
      \includegraphics[width=0.13\textwidth]{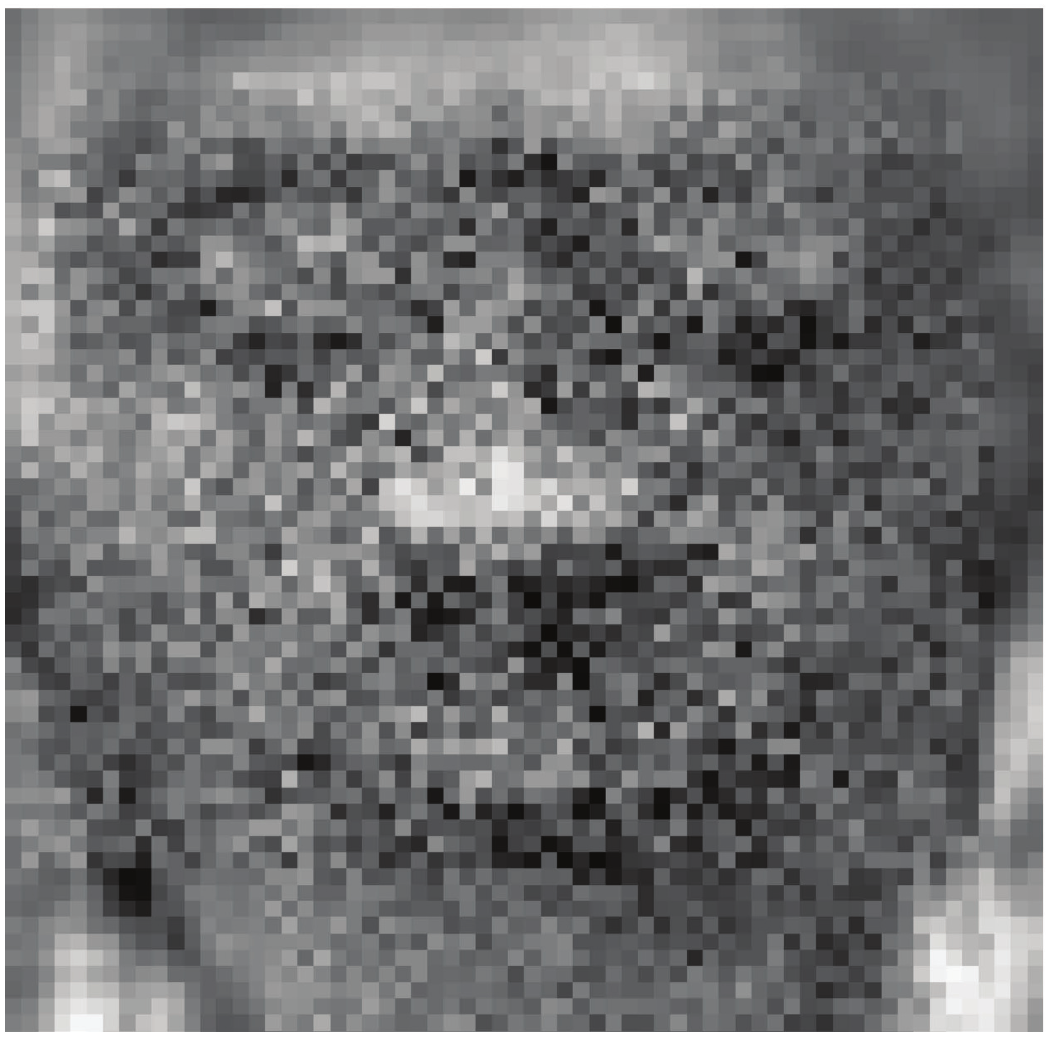}
      \includegraphics[width=0.13\textwidth]{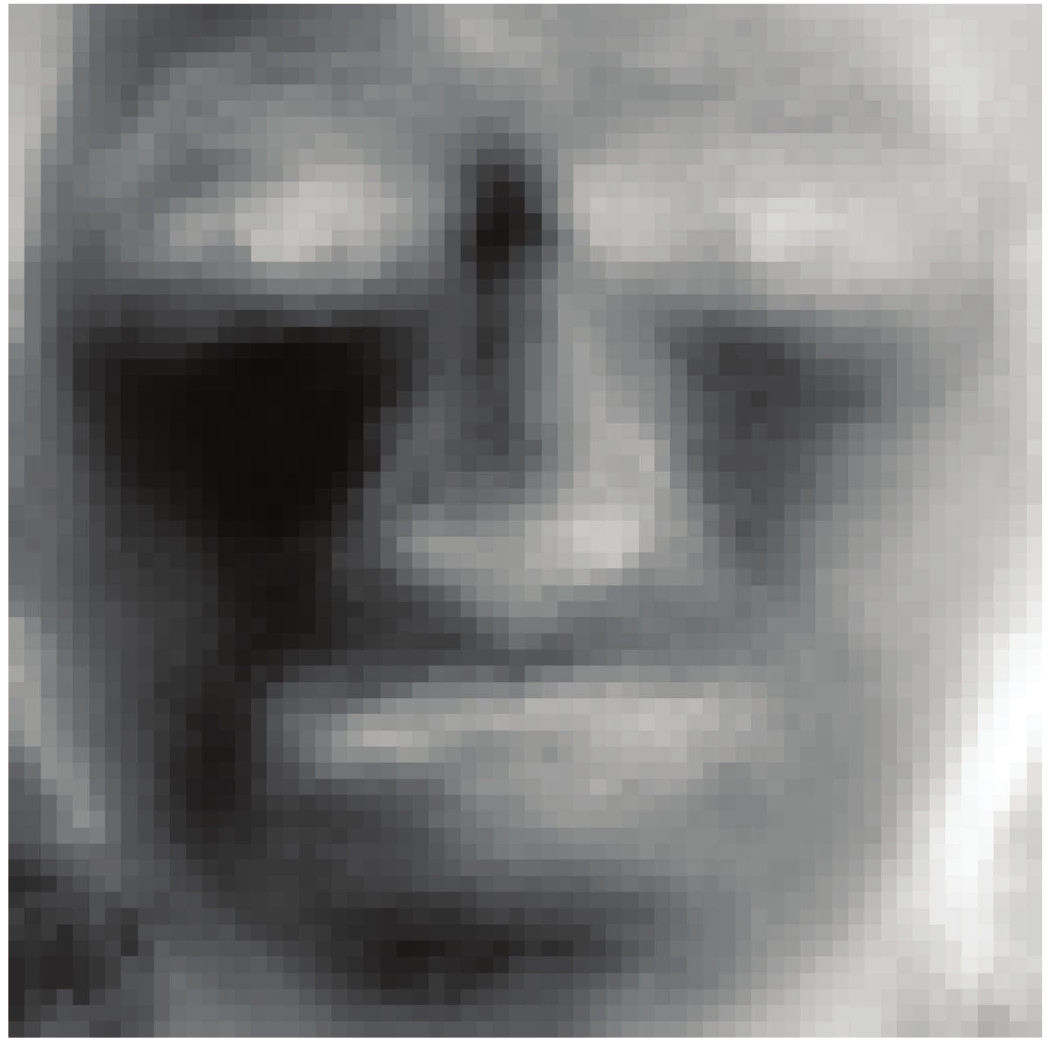}
      \includegraphics[width=0.13\textwidth]{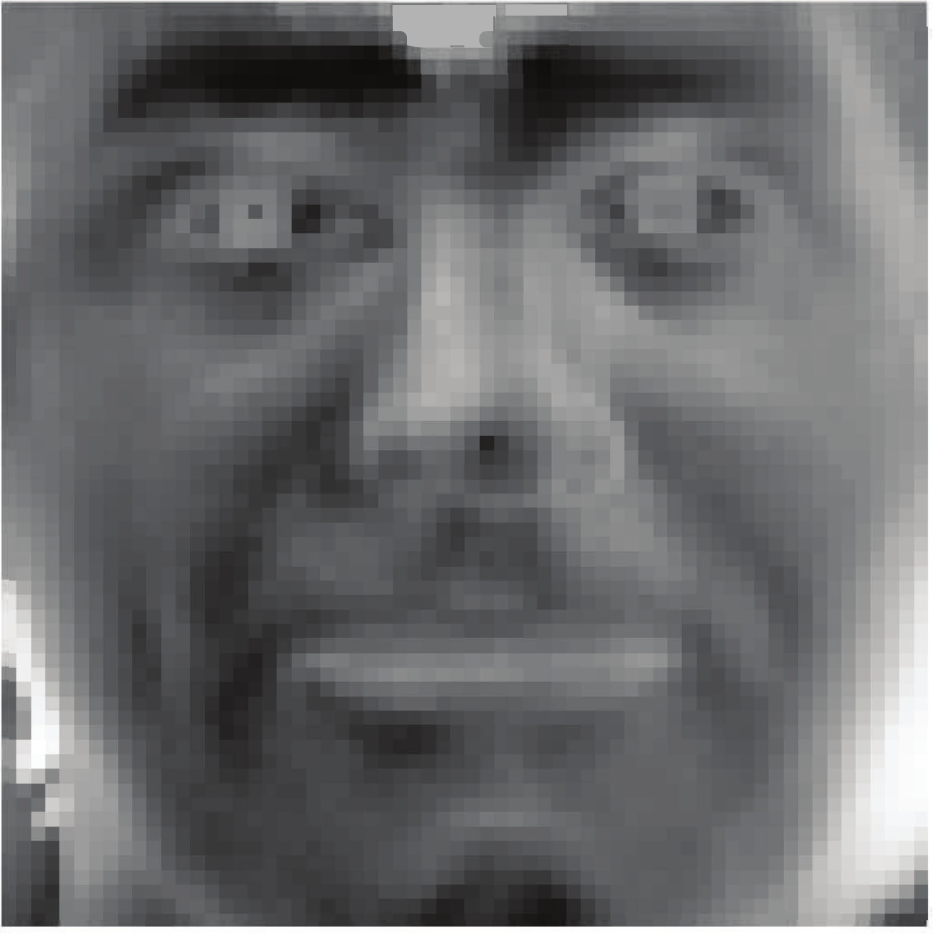}
      \includegraphics[width=0.13\textwidth]{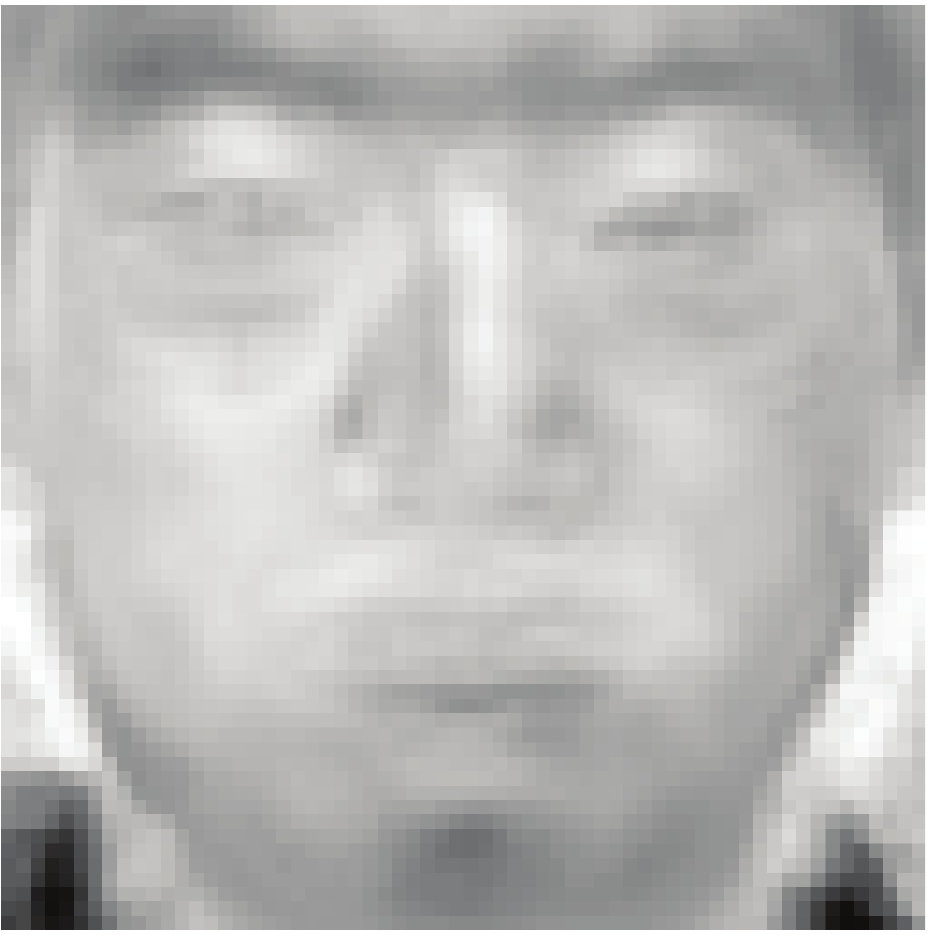}
      \includegraphics[width=0.13\textwidth]{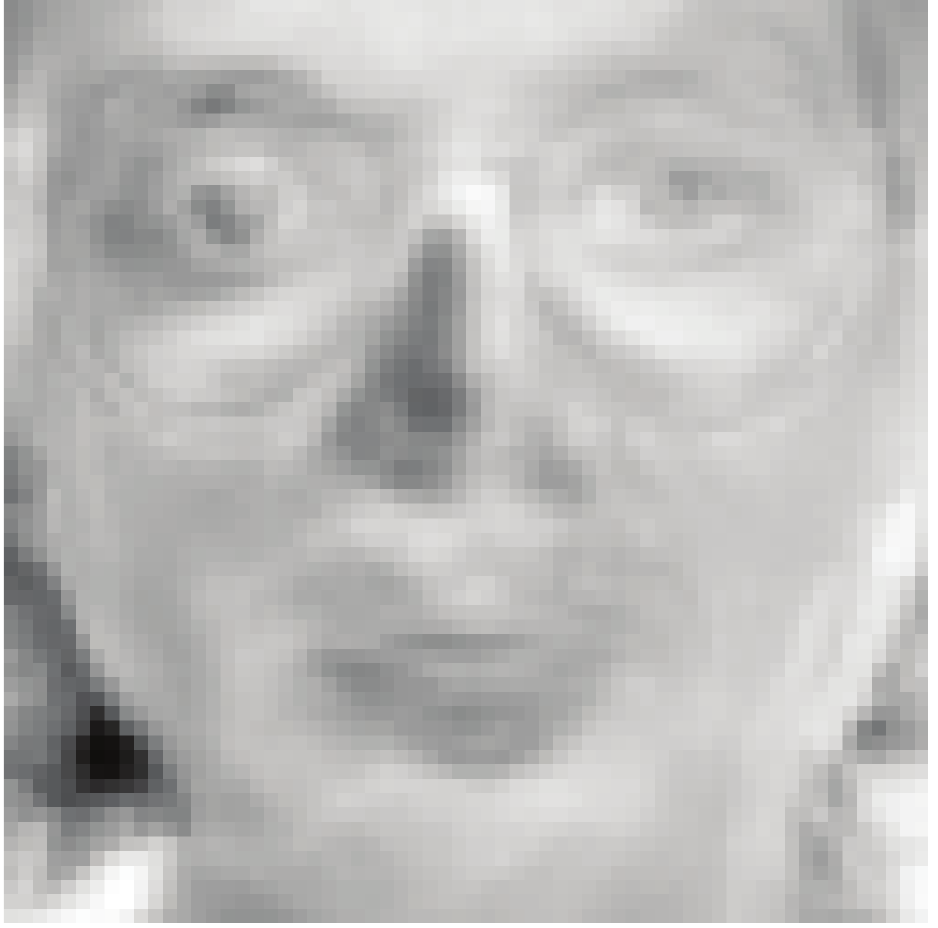}\\
      \includegraphics[width=0.13\textwidth]{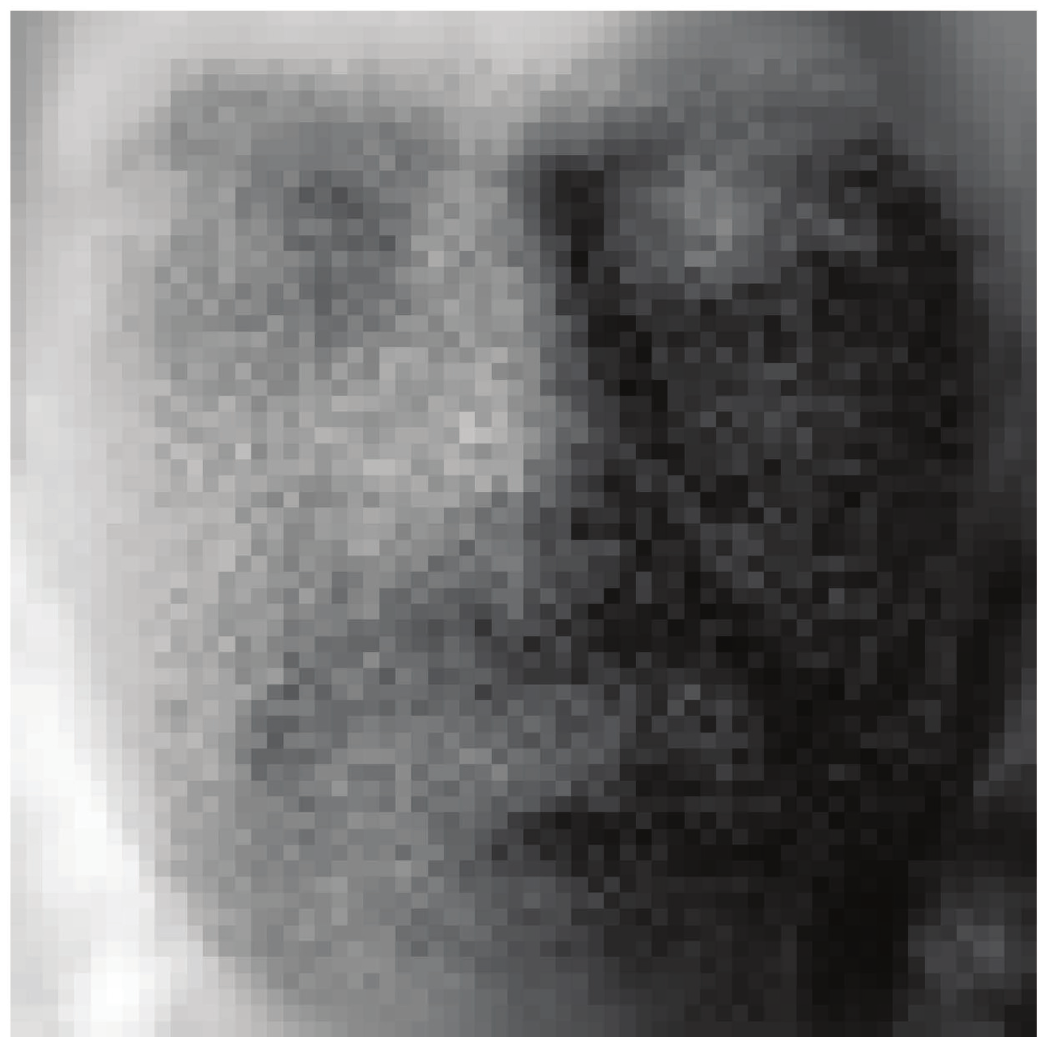}
      \includegraphics[width=0.13\textwidth]{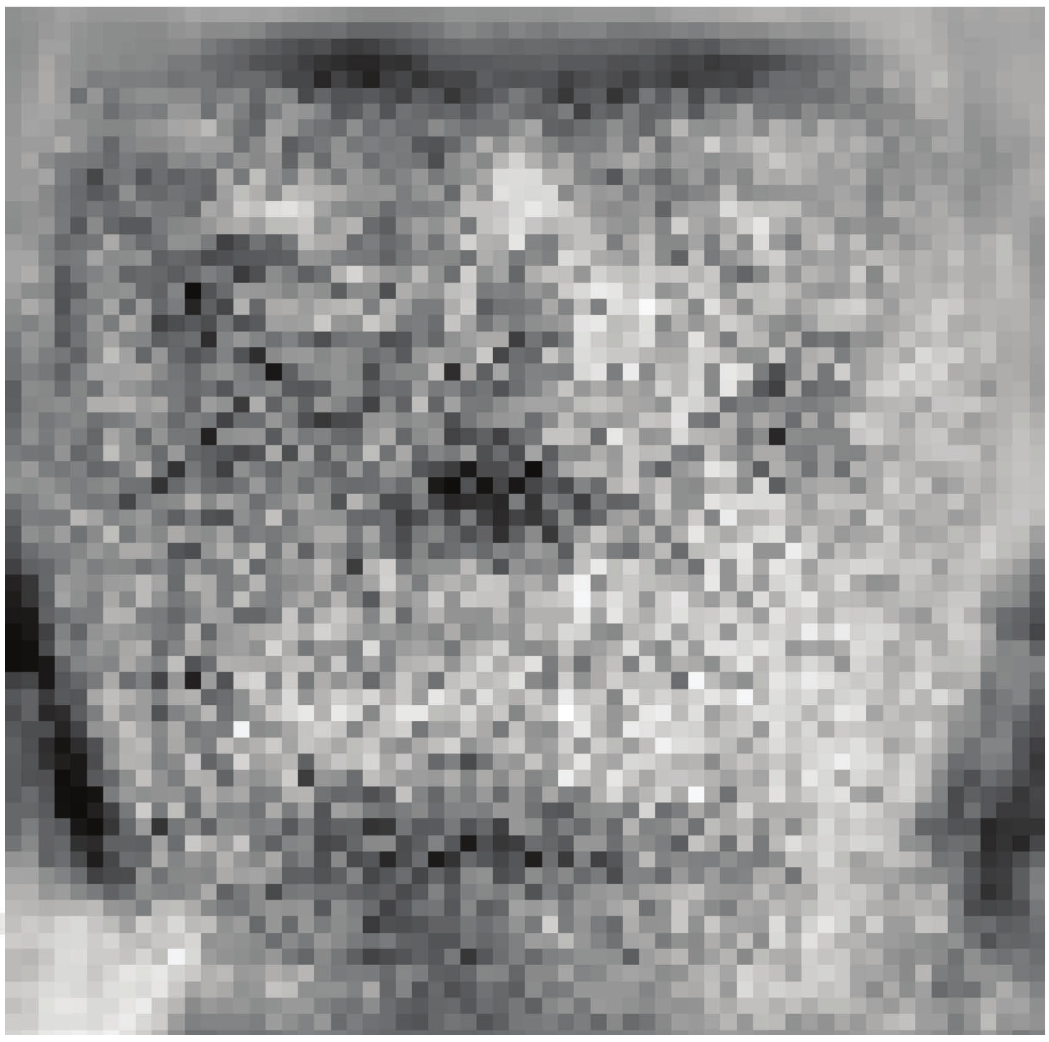}
      \includegraphics[width=0.13\textwidth]{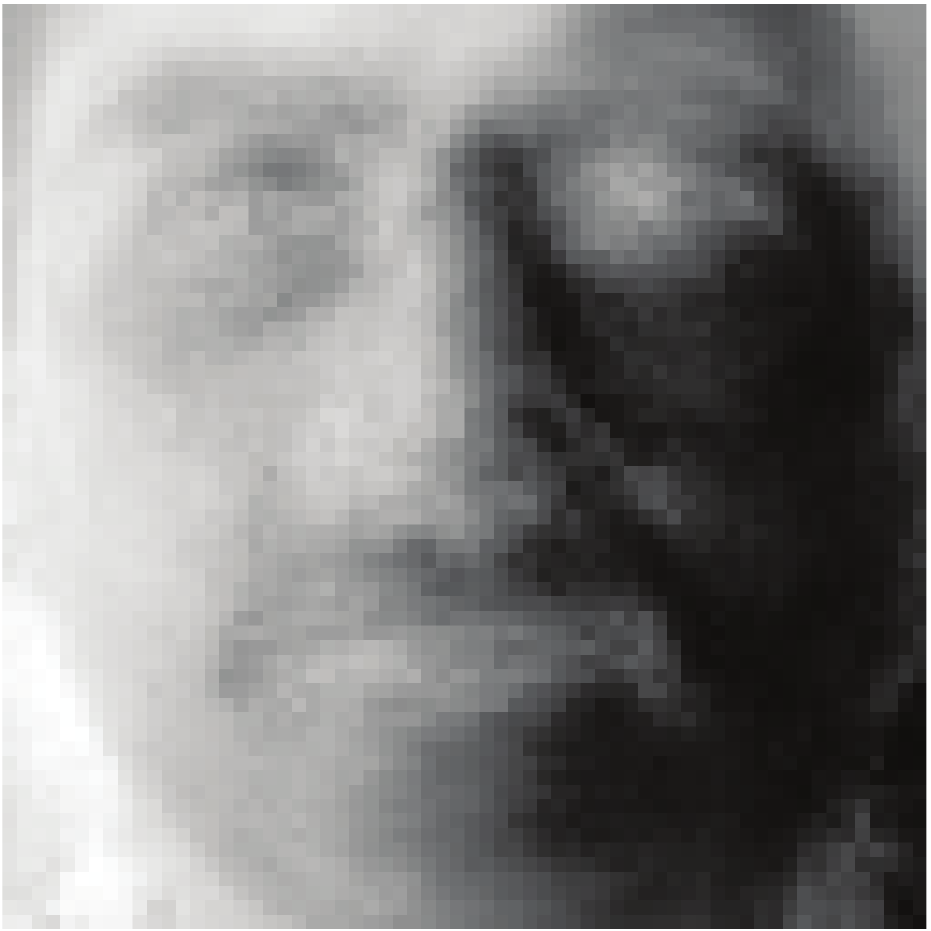}
      \includegraphics[width=0.13\textwidth]{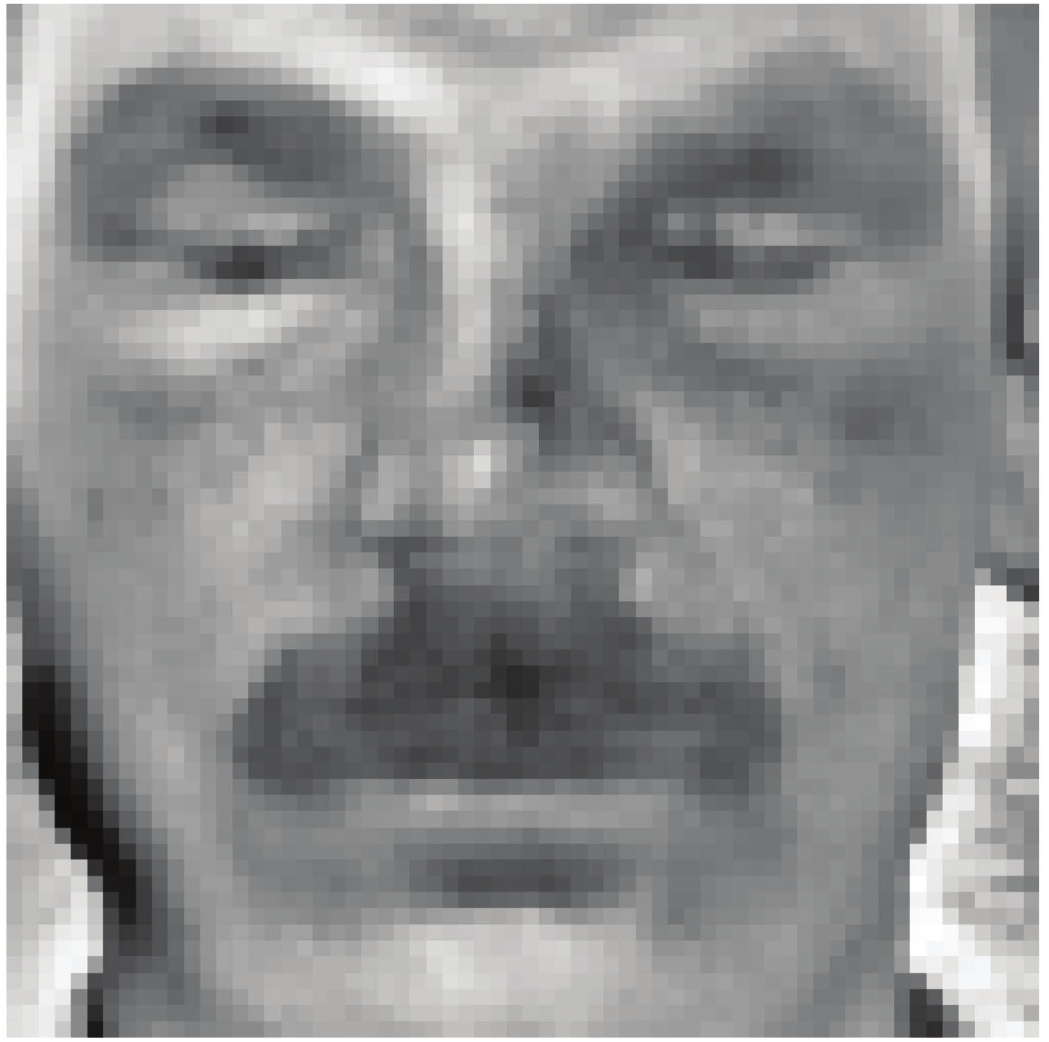}
      \includegraphics[width=0.13\textwidth]{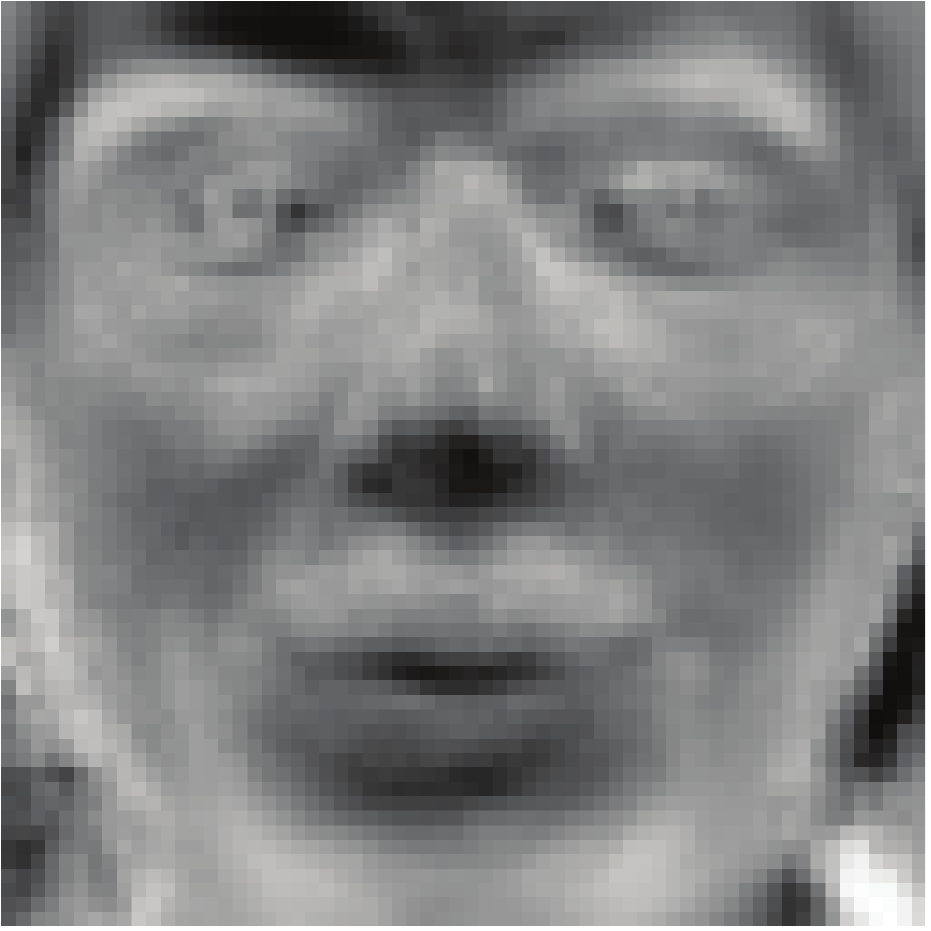}
      \includegraphics[width=0.13\textwidth]{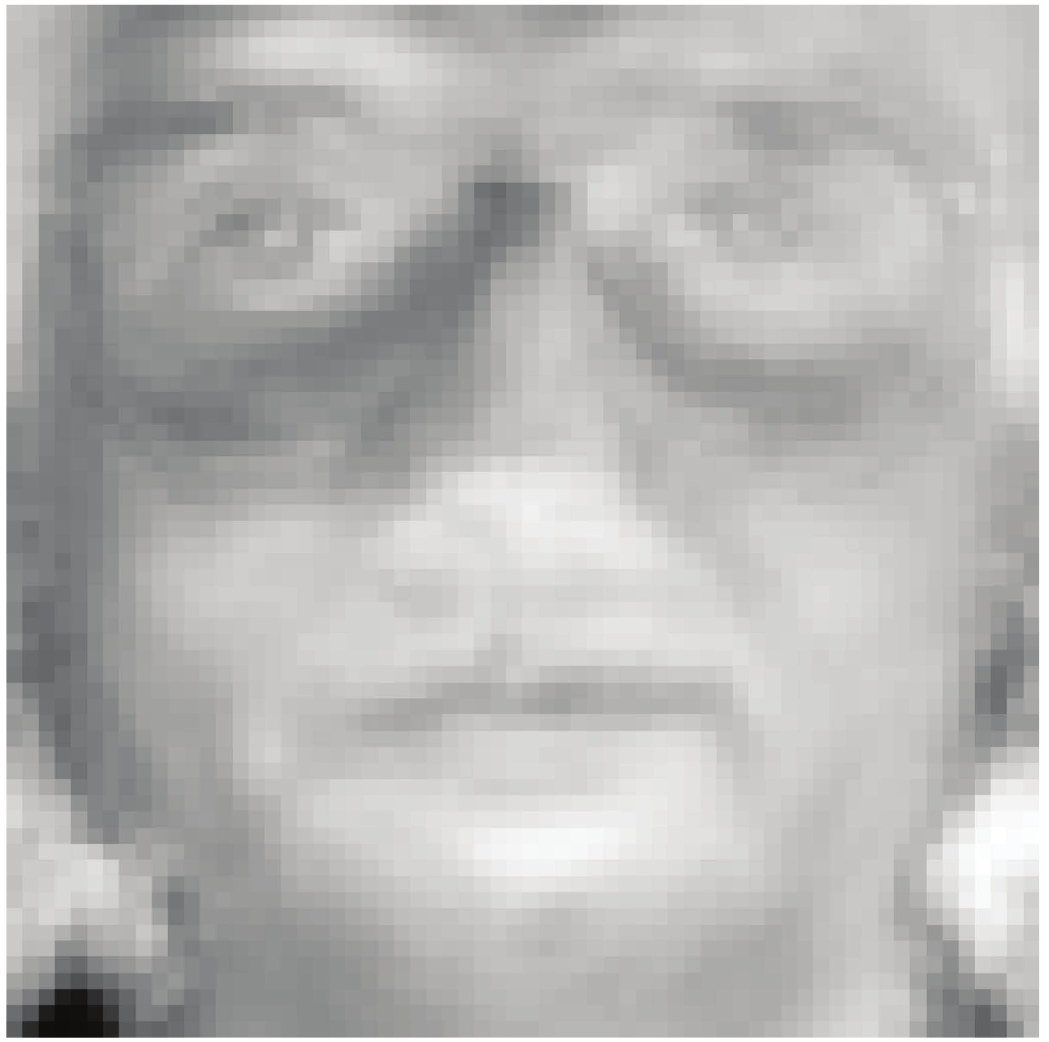}\\
      \includegraphics[width=0.13\textwidth]{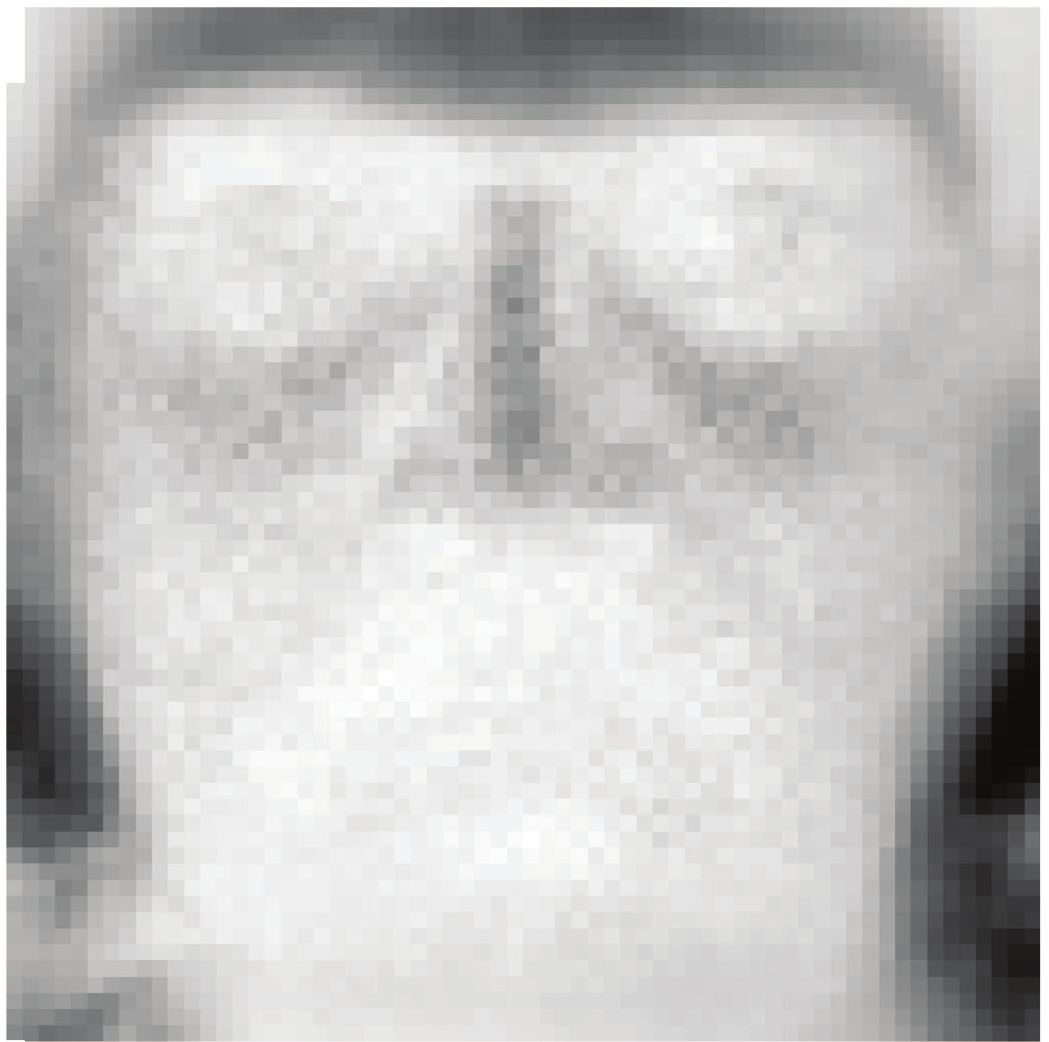}
      \includegraphics[width=0.13\textwidth]{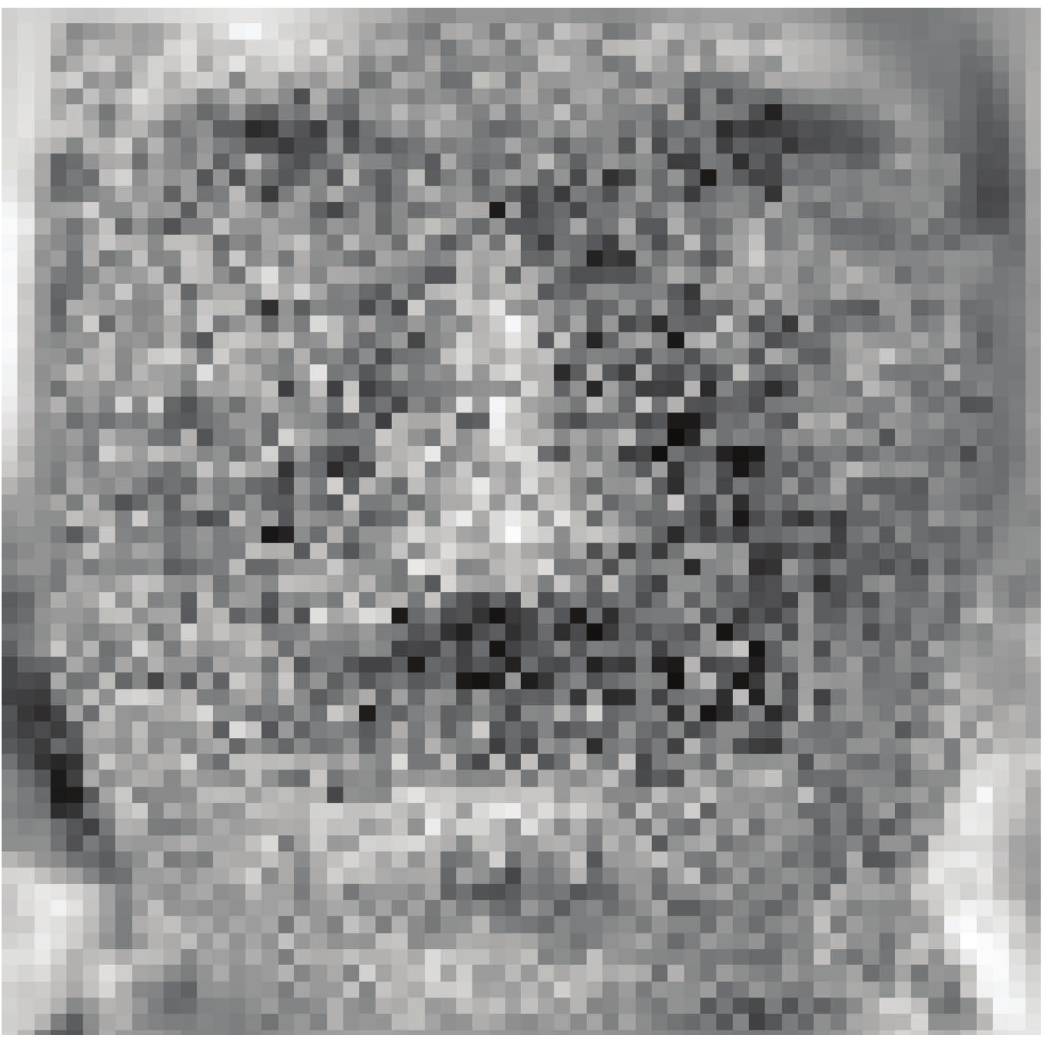}
      \includegraphics[width=0.13\textwidth]{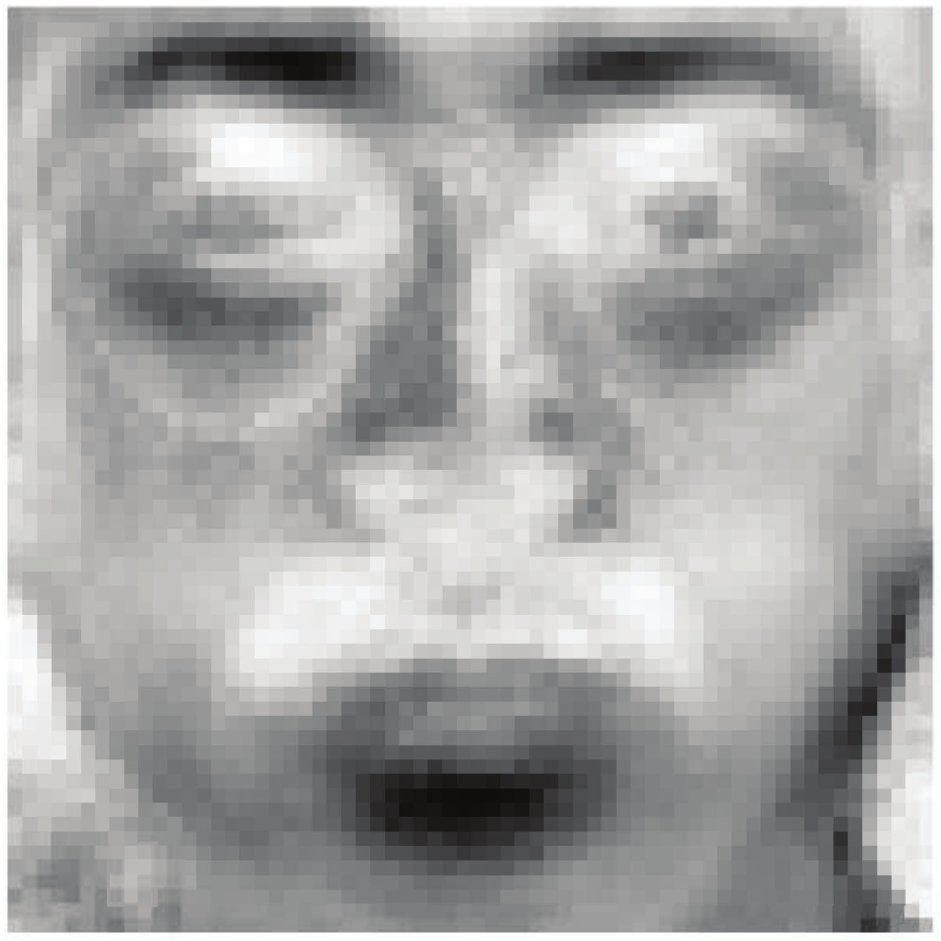}
      \includegraphics[width=0.13\textwidth]{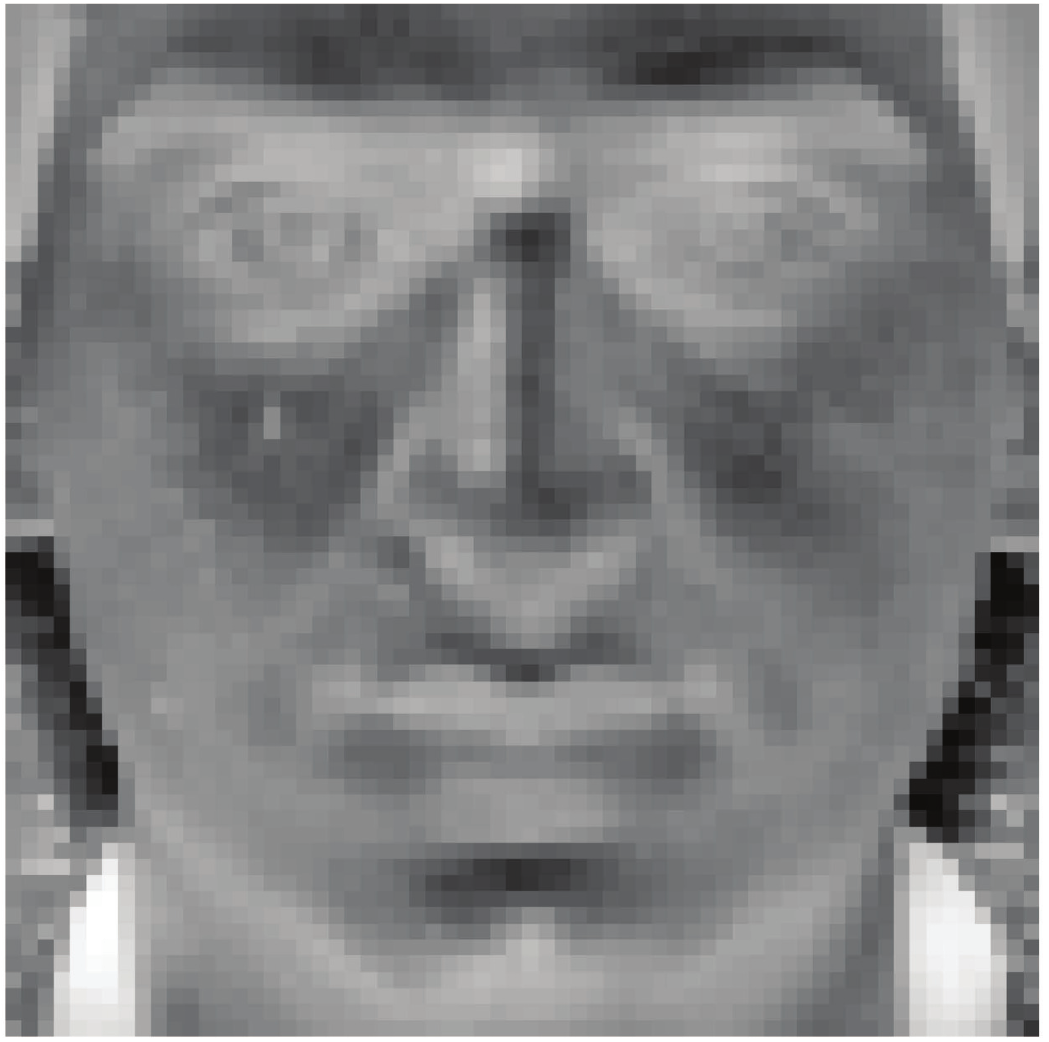}
      \includegraphics[width=0.13\textwidth]{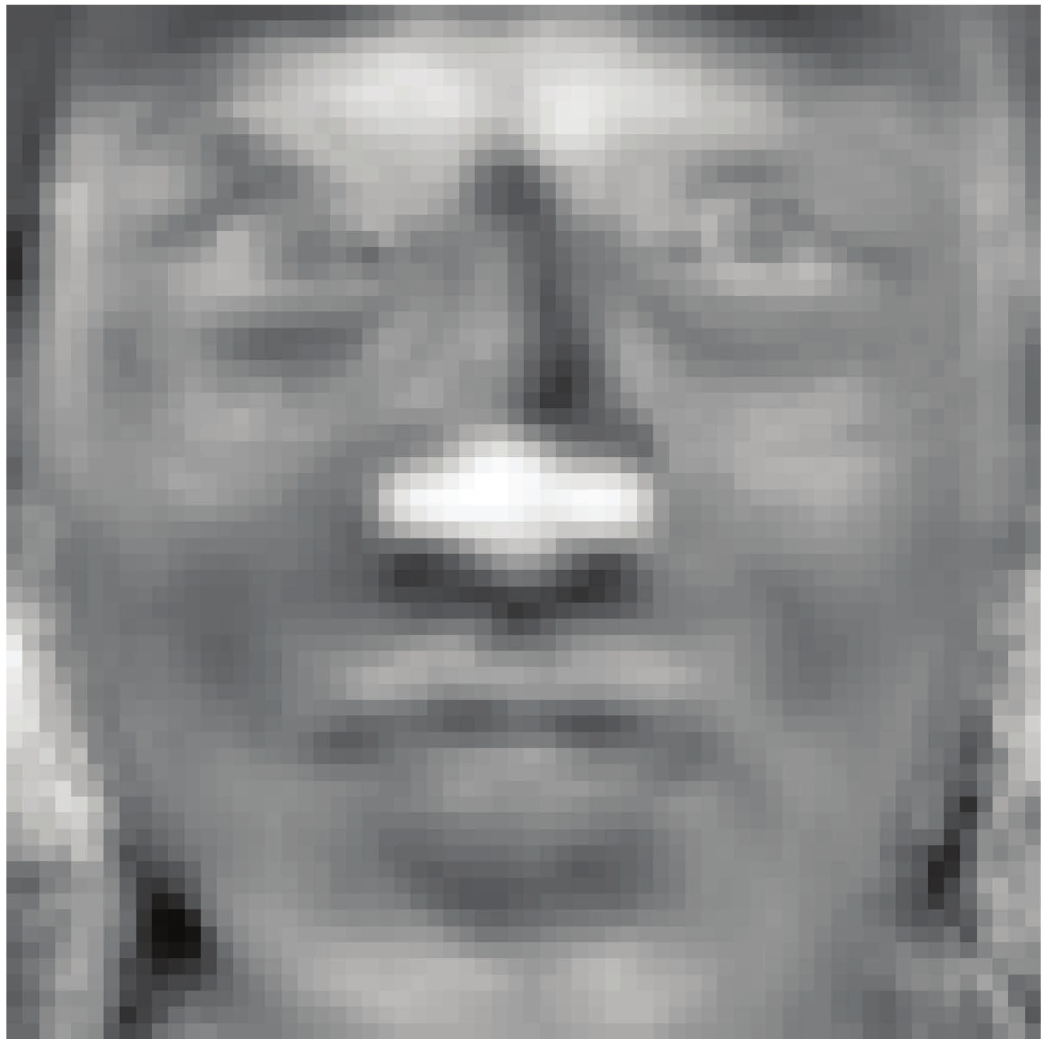}
      \includegraphics[width=0.13\textwidth]{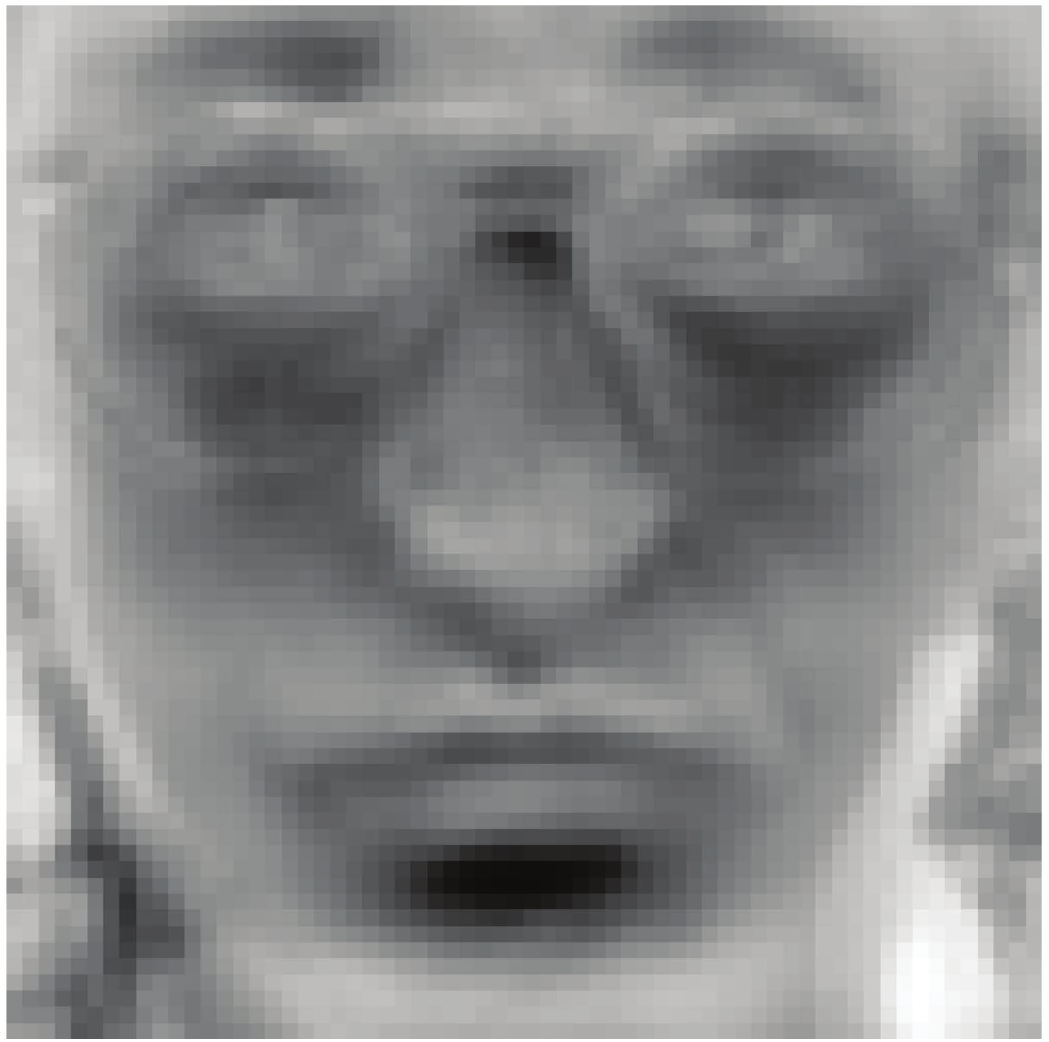}\\
      \includegraphics[width=0.13\textwidth]{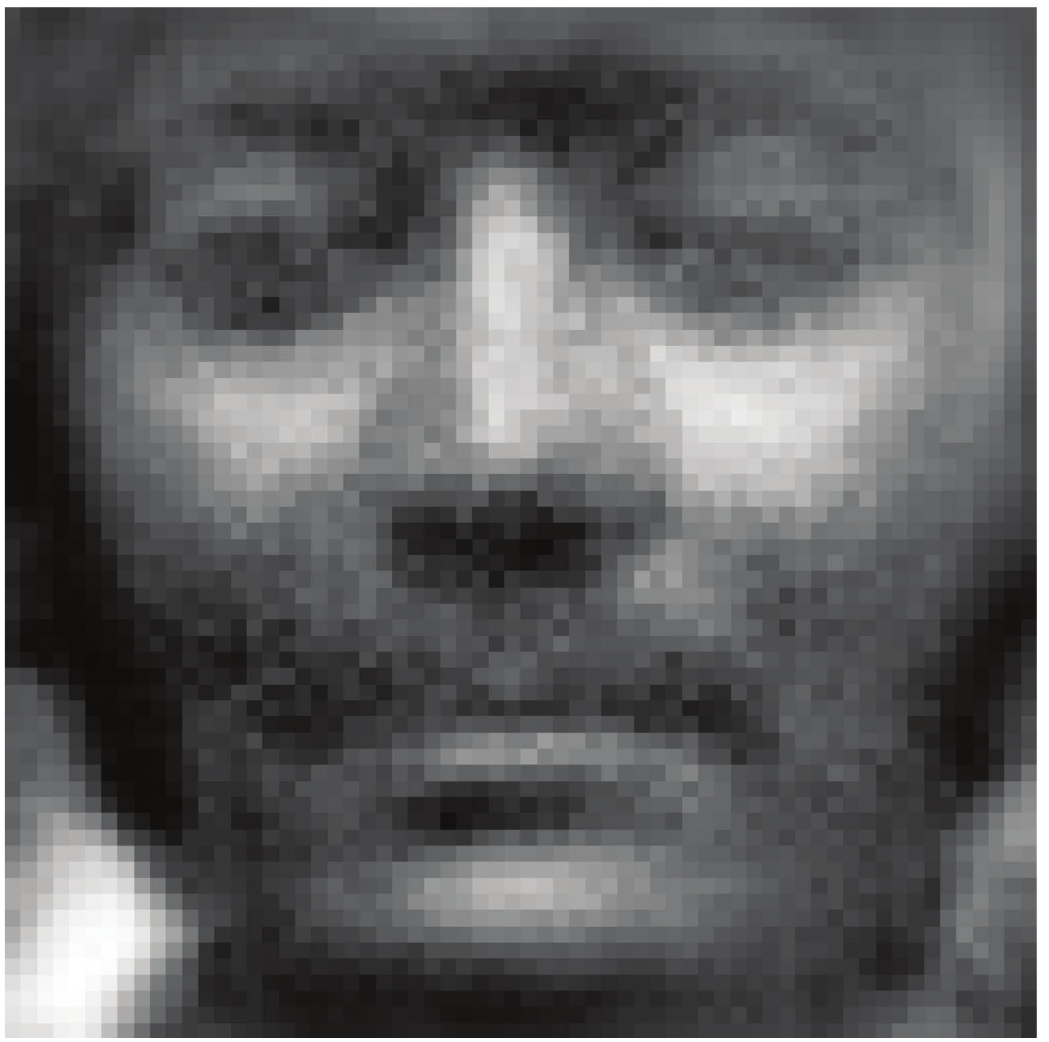}
      \includegraphics[width=0.13\textwidth]{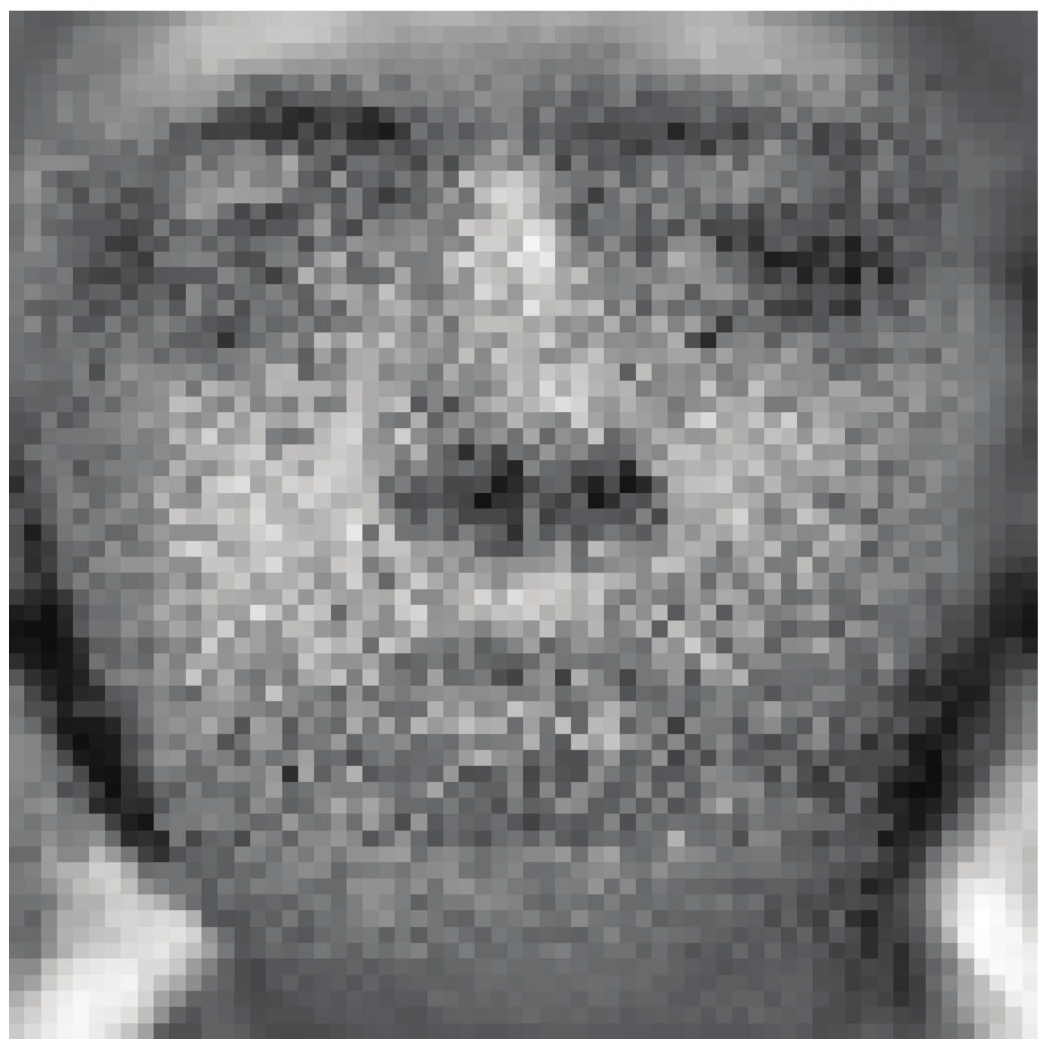}
      \includegraphics[width=0.13\textwidth]{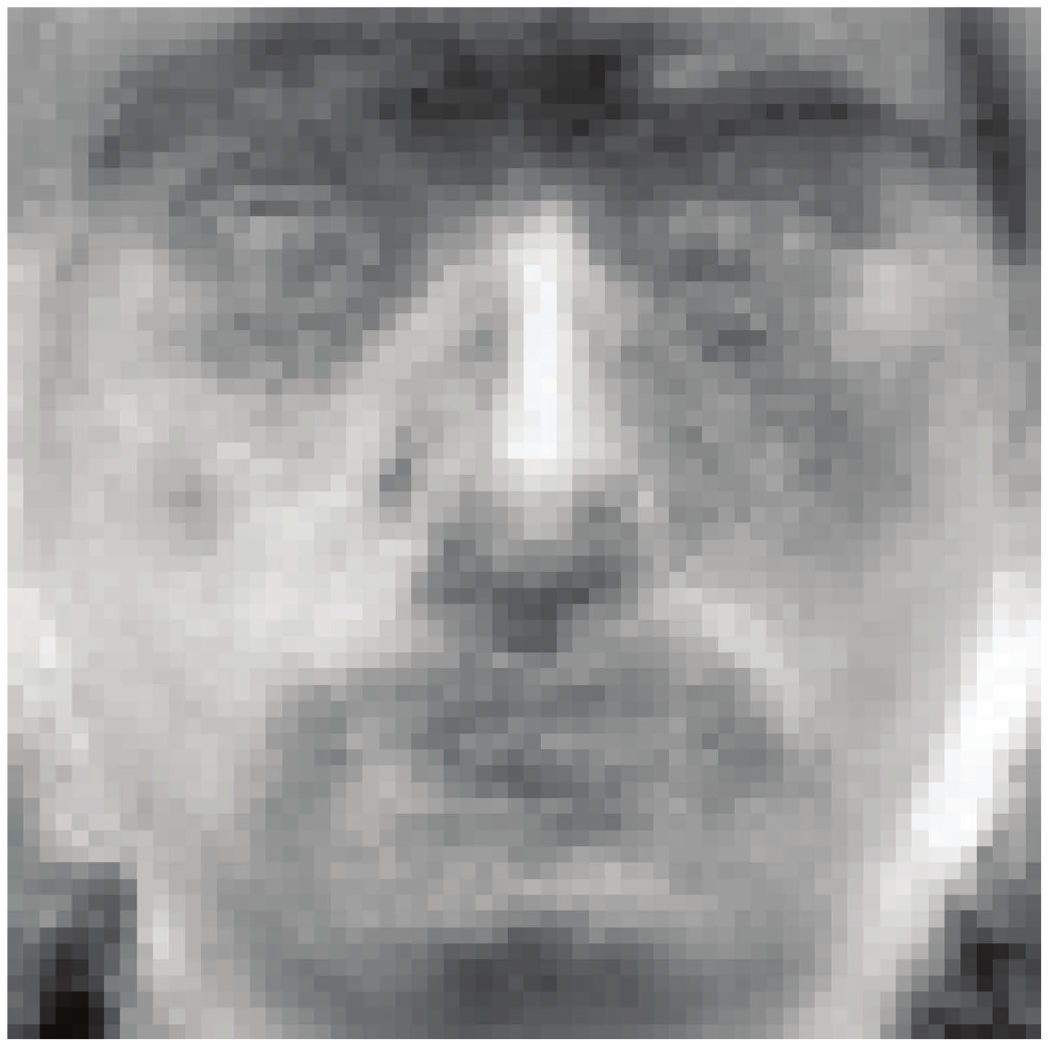}
      \includegraphics[width=0.13\textwidth]{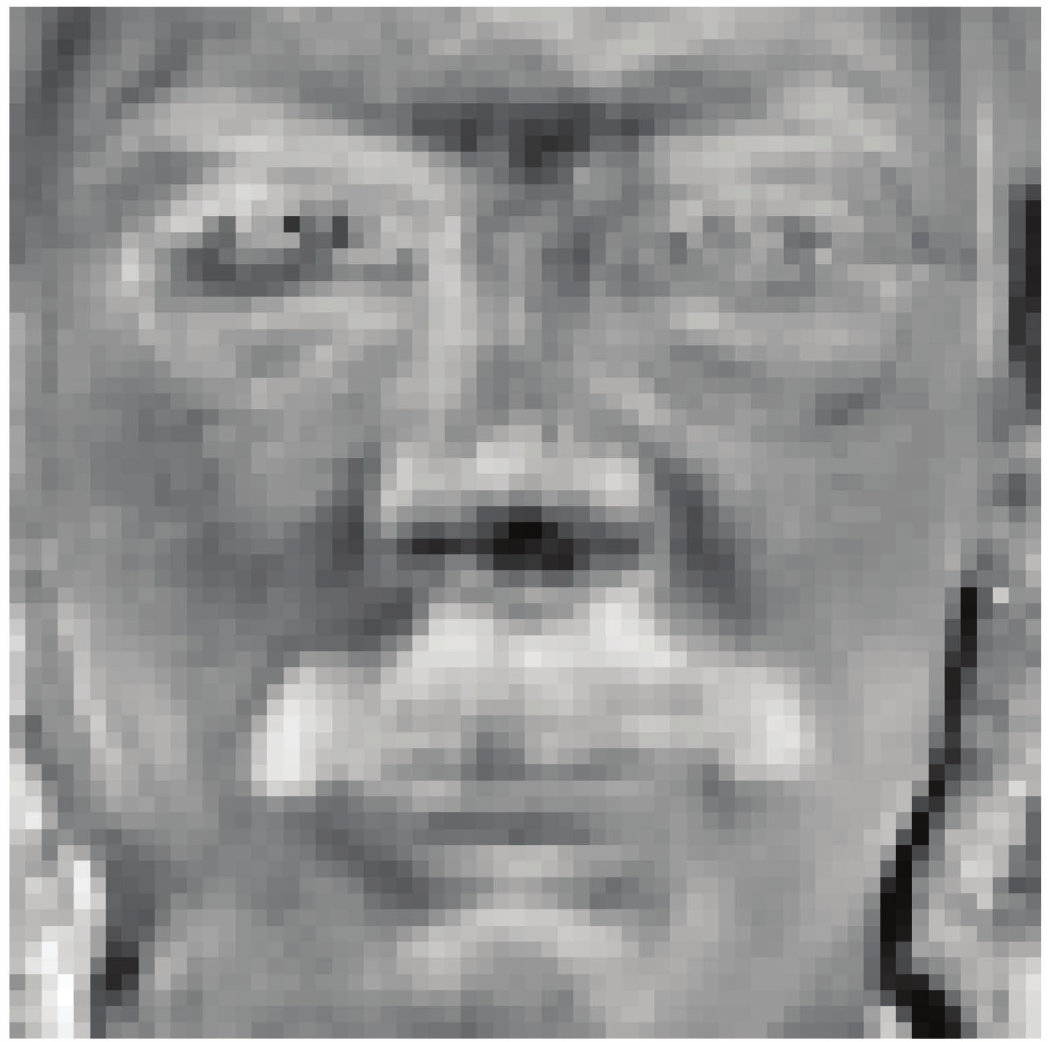}
      \includegraphics[width=0.13\textwidth]{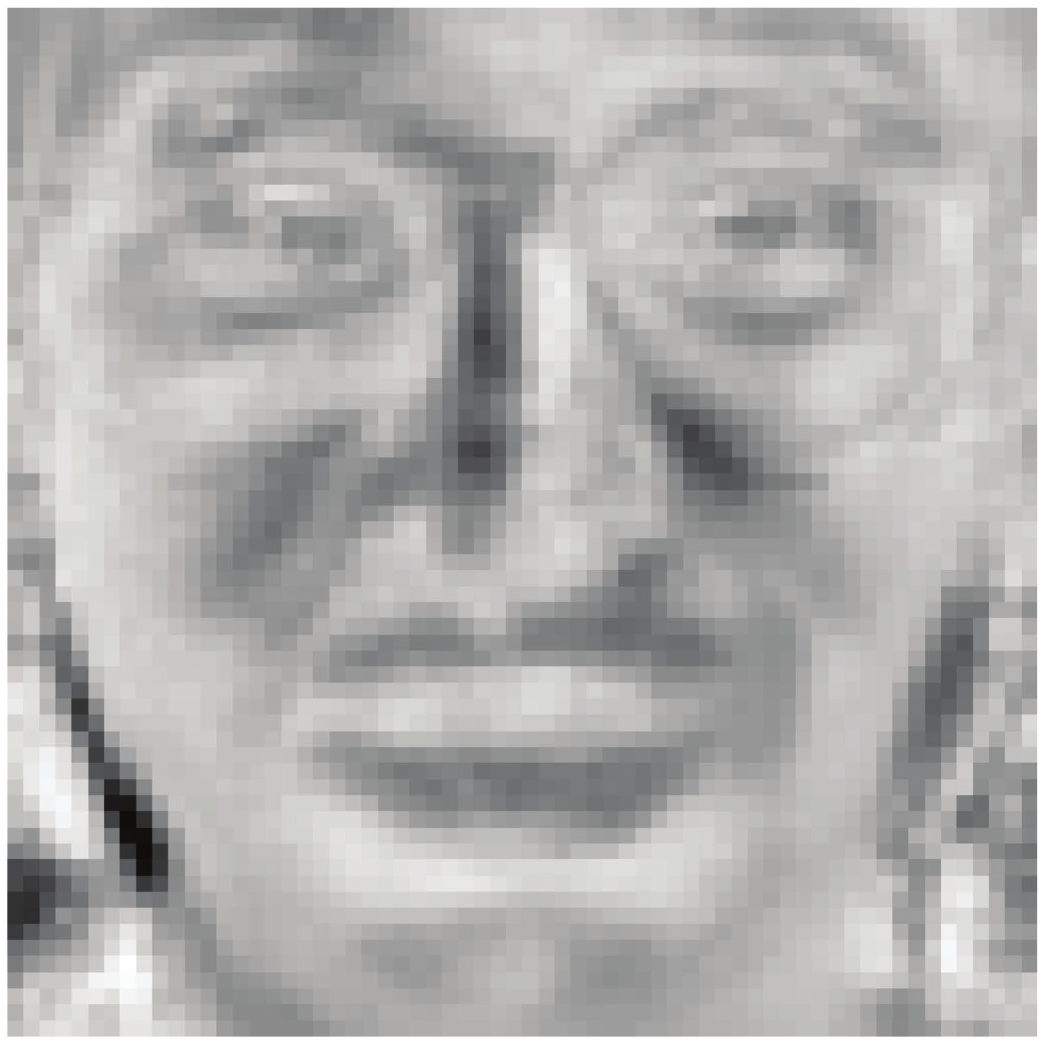}
      \includegraphics[width=0.13\textwidth]{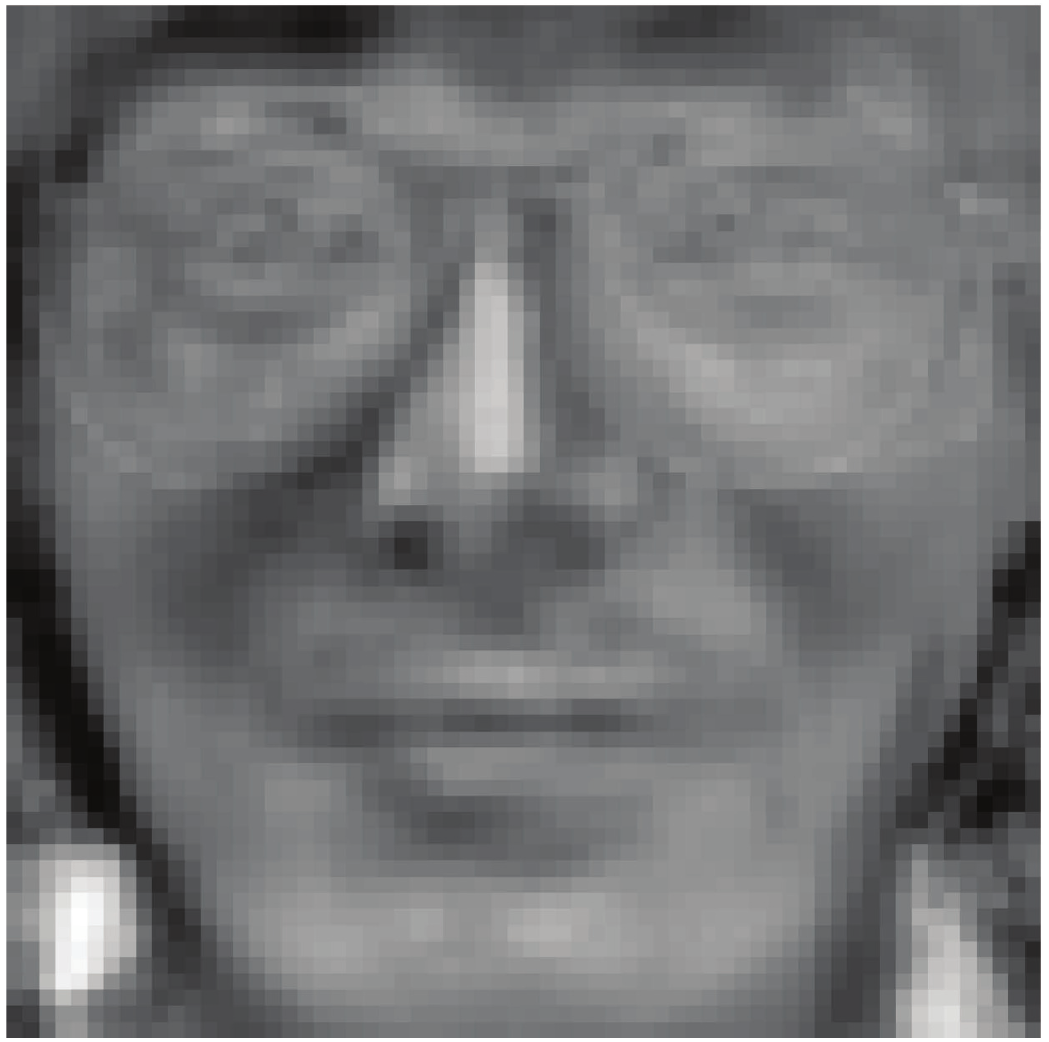}
      \caption{Some eigenfaces obtained by different methods. The first column shows eigenfaces of PCA, the second column PPCA, the third column PCP, the fourth column RAPCA, the fifth column ROBPCA, and the last column SP-PPCA (our method).}
      \label{fig_eigenface}
\end{figure}

\begin{figure}[htbp]
      \centering
      \includegraphics[width=0.98\textwidth]{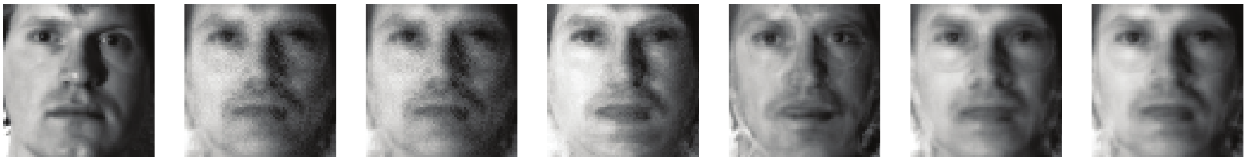}\\
      \includegraphics[width=0.98\textwidth]{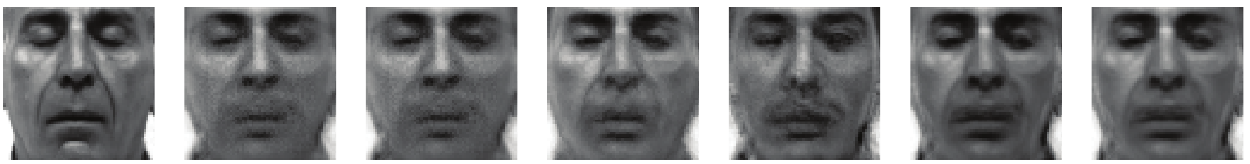}\\
      \includegraphics[width=0.98\textwidth]{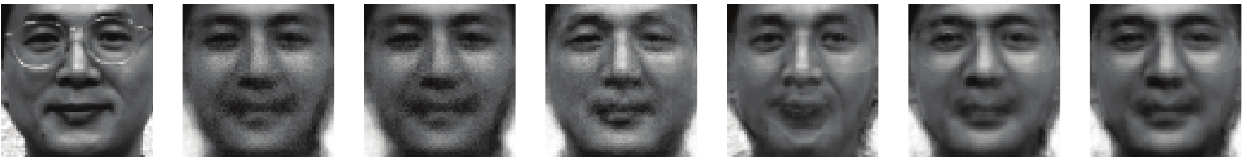}\\
      \includegraphics[width=0.98\textwidth]{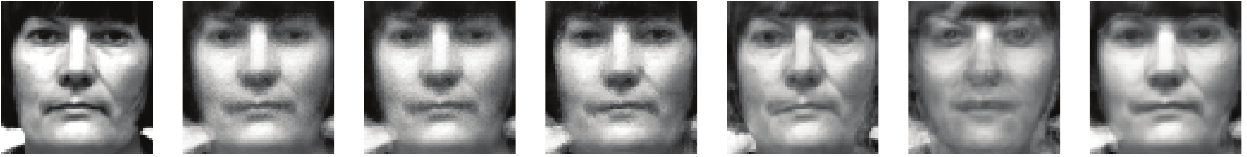}\\
      \includegraphics[width=0.98\textwidth]{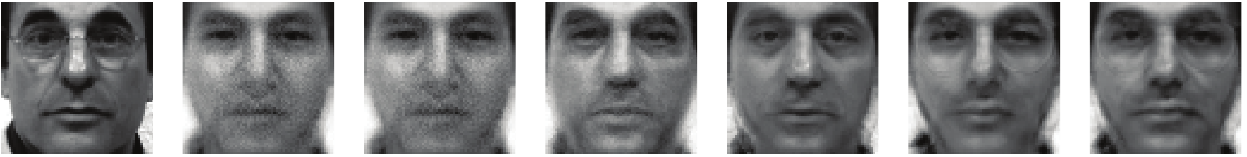}\\
      \caption{Reconstructed images by six different methods. The first column shows images in the test set, the second column are reconstructed images by PCA, the third column by PPCA, the fourth column by PCP, the fifth column by RAPCA, the sixth column by ROBPCA and the last column by SP-PPCA (our method).}
      \label{fig_face_recover}
\end{figure}

Figure \ref{fig_eigenface} compares the eigenfaces of six different methods. As we can see, the eigenfaces of most methods are polluted by the outlying images in the training set. Compared with other approaches, the eigenfaces of SP-PPCA are less affected by the contaminated data. Figure \ref{fig_face_recover} presents some images in the test set and the corresponding reconstructed images using 30 projection vectors. We can see that the images reconstructed by PCA, PPCA, PCP and RAPCA are impacted by outliers to some degree. SP-PPCA and ROBPCA perform better than other methods. The success of SP-PPCA is due to Self-Paced Learning mechanism which tends to filter out outliers in the training set. Therefore, the projection vectors computed from SP-PPCA are less influenced by the outlying images. Table \ref{tab_faceresult} further provides the average reconstruction errors of each method. It can be shown that in most cases, the results of SP-PPCA are better than those of other methods.

\renewcommand\arraystretch{1.07}
\begin{table}[htbp]
      \centering
      \caption{Average reconstruction errors of different methods. ``$M$'' is the dimension of the reduced subspace; ``num \& size'' represent the number of outliers in the training dataset and the size of noise blocks. The best results are highlighted in bold.}
      \setlength{\tabcolsep}{2mm}{
        \begin{tabular}{ccccccccc}
        \toprule
        \multicolumn{2}{c}{num \& size} & $M$     & PCA   & PPCA  & PCP   & RAPCA & ROBPCA & SP-PPCA \\
        \midrule
        \multicolumn{2}{c}{15 \&} & 20    & 0.2108  & 0.2107  & 0.2106  & 0.2424  & 0.2170  & \textbf{0.2025 } \\
        \multicolumn{2}{c}{$30\times 30$} & 30    & 0.2027  & 0.2026  & 0.2059  & 0.2221  & 0.2094  & \textbf{0.1963 } \\
        \multicolumn{2}{c}{} & 40    & 0.1862  & 0.1860  & 0.1901  & 0.1974  & 0.1886  & \textbf{0.1786 } \\
        \midrule
        \multicolumn{2}{c}{30 \&} & 20    & 0.2140  & 0.2140  & 0.2133  & 0.2481  & 0.2232  & \textbf{0.2099 } \\
        \multicolumn{2}{c}{$30\times 30$} & 30    & 0.2015  & 0.2015  & 0.1985  & 0.2176  & 0.2081  & \textbf{0.1931 } \\
        \multicolumn{2}{c}{} & 40    & 0.1976  & 0.1976  & \textbf{0.1922 } & 0.2051  & 0.2015  & 0.1940  \\
        \midrule
        \multicolumn{2}{c}{15 \&} & 20    & 0.2262  & 0.2262  & 0.2208  & 0.2488  & 0.2272  & \textbf{0.2148 } \\
        \multicolumn{2}{c}{$45\times 45$} & 30    & 0.2101  & 0.2101  & 0.2001  & 0.2209  & 0.1998  & \textbf{0.1875 } \\
        \multicolumn{2}{c}{} & 40    & 0.1922  & 0.1921  & 0.1918  & 0.2011  & 0.1856  & \textbf{0.1762 } \\
        \bottomrule
        \end{tabular}}%
      \label{tab_faceresult}%
\end{table}%

\section{Related work}
\label{relatedwork}
\vspace{-0.5mm}
Several robust PCA approaches have been proposed in the literature to handle outliers. Here we mainly introduce three kinds of methods: $L_1$-norm based PCA,  Projection Pursuit based PCA and Principal Component Pursuit method.

PCA is based on the $L_2$-norm, which is not robust to outliers, so that much of the research work resorts to the  $L_1$-norm to counteract outliers \cite{baccini1996l1, ke2005robust, dhanaraj2018novel, nie2011robust, ju2015image}. Baccini et al. \cite{baccini1996l1} introduced Gini's mean absolute differences into PCA to get heuristic estimates for the $L_1$-norm PCA model. Through using alternative convex programming, a subspace estimation method which minimizes the $L_1$-norm based cost function was presented in \cite{ke2005robust}. In addition, Markopoulos et al. \cite{markopoulos2017efficient} proposed the L1-BF algorithm to efficiently calculate the principal components of the $L_1$-norm PCA based on bit-flipping. However, outliers still persist in the training process of $L_1$-norm PCA methods, so these approaches can only alleviate the impact of outliers to some extent. Moreover, current $L_1$-norm PCA methods have a high computational cost \cite{minnehan2019grassmann}. 

Projection Pursuit \cite{li1985projection} based robust PCA attempts to search for the direction that maximize a robust measure of the variance of the projected observations. 
In order to relieve the high computational cost problem in \cite{li1985projection}, a faster algorithm, named RAPCA, was presented in \cite{hubert2002fast}.
Furthermore, Hubert et al. \cite{hubert2005robpca} combined projection pursuit ideas with robust covariance estimator and proposed the method ROBPCA. 

In Principal Component Pursuit method (PCP) \cite{candes2011robust}, the observed data is decomposed into a low rank matrix and a sparse matrix, among which the sparse matrix can be treated as noise. While, such a decomposition is not always realistic \cite{neumayer2019rotational}. Moreover, PCP is proposed for uniformly corrupted coordinates of data, rather than outlying samples \cite{zhang2014novel}.

In addition, based on statistical physics and self-organized rules, a robust PCA was proposed in \cite{xu1995robust}. The authors generalized an energy function by introducing binary variables. We also adopt binary variables in our algorithm, however, our method is derived under the probability framework based on Self-Paced Learning and Probabilistic PCA. More importantly, we give an efficient strategy to solve the relevant optimization problem.

\vspace{-3mm}
\section{Conclusion}
\label{sec5}
\vspace{-0.5mm}
In this paper, with the aim to build a more robust dimensionality reduction algorithm, we propose a novel model called SP-PPCA. To our knowledge, this is the first attempt to introduce Self-Paced Learning mechanism into Probabilistic PCA. We develop the corresponding algorithm for optimizing the model parameters. Compared with classical PCA and several variant algorithms, extensive experiments on the simulation data and real-world data demonstrate the effectiveness of our model. 

SP-PPCA can be easily extended to solve the missing value problem like PPCA. In addition, SP-PPCA is based on Gaussian distribution, other robust distribution, such as heavy-tailed distributions \cite{luttinen2012bayesian}, can be adopted in SP-PPCA to improve the robustness in the future.

%
%
%
\bibliographystyle{splncs04}
\bibliography{mybibliography}

\end{document}